\documentclass[11pt]{article} 
\pdfoutput=1
\usepackage{geometry} 
\usepackage[utf8]{inputenc}
\usepackage{caption}
\usepackage{subcaption}
\usepackage{graphicx} 

\usepackage[nottoc,notlof,notlot]{tocbibind} 
\usepackage[square,numbers]{natbib}
\usepackage[titles,subfigure]{tocloft} 
\usepackage{booktabs} 
\usepackage{array} 
\usepackage{paralist} 
\usepackage{verbatim} 

\usepackage{amssymb, amsmath, amsthm,tikz,listings,esint,cancel,mathabx, mathtools} 

\usepackage{setspace}

\usepackage{hyperref}
\hypersetup{colorlinks,allcolors=black}

\usepackage{filecontents}

\begin{document}

\begin{titlepage}
   \begin{center}
       \vspace*{1cm}
        \Huge
       \textbf{Clustering US Counties to Find Patterns Related to the COVID-19 Pandemic}
\Large     
       \vspace{1.5cm}
       
       \textbf{Cora Brown\\Sarah Milstein\\Tianyi Sun\\Cooper Zhao}
       \vfill         
       \vspace{0.8cm}            
       University of Minnesota, School of Mathematics\\
       in collaboration with Ecolab\\
       November 16, 2020
       
   \end{center}
\end{titlepage}

\tableofcontents
\newpage
\section{Introduction}
When COVID-19 first started spreading and quarantine was implemented, the Society for Industrial and Applied Mathematics (SIAM) Student Chapter at the University of Minnesota-Twin Cities began a collaboration with Ecolab to use our skills as data scientists and mathematicians to extract useful insights from relevant data relating to the pandemic. This collaboration consisted of multiple groups working on different projects. In this write-up we focus on using clustering techniques to help us find groups of similar counties in the US and use that to help us understand the pandemic. Our team for this project consisted of University of Minnesota students Cora Brown, Sarah Milstein, Tianyi Sun, and Cooper Zhao, with help from Ecolab Data Scientist Jimmy Broomfield and University of Minnesota student Skye Ke.

In the sections below we describe all of the work done for this project. In Section 2, we list the data we gathered, as well as the feature engineering we performed. In Section 3, we describe the metrics we used for evaluating our models. In Section 4, we explain the methods we used for interpreting the results of our various clustering approaches. In Section 5, we describe the different clustering methods we implemented. In Section 6, we present the results of our clustering techniques and provide relevant interpretation. Finally, in Section 7, we provide some concluding remarks comparing the different clustering methods.

\section{Data}
For this project, we used data from a variety of sources, including the collection of data for the UNCOVER COVID-19 Kaggle Challenge\footnote{Available at \url{https://www.kaggle.com/roche-data-science-coalition/uncover}} and case and death numbers from the Centers for Disease Control. 

Out of the available data for the UNCOVER Challenge, we chose and engineered features which could be relevant in COVID-19 clustering. Here we group our final set of features into several groups in order to more easily describe their relevance to the problem at hand. First, we have features which represent the makeup and demographics of each county, including 
\begin{enumerate}
\item Area - square miles of area in each county,

\item Population - population of each county,

\item Ranking: socioeconomic status - To determine this ranking, the CDC first computed the percentile ranking (0 to 1 with 1 meaning greater vulnerability) for the following features, based on Census data, that they determined to be in the category of socioeconomic status: (1) percentage of people below the poverty line, (2) percentage of people, age 16 and older, who are unemployed, (3) per capita income, and (4) percentage of people, age 25 and older, without a high school diploma. They then summed together the percentile ranking across all 4 features for each county, ranked these resulting sums, and computed each county's percentile ranking with respect to this sum as the final column value.

\item Ranking: household decomposition and disability - Computed following the same procedure as `ranking: socioeconomic status' except the features used were: (1) percentage of people aged 65 and older, (2) percentage of people aged 17 and younger, (3) percentage of civilians with a disability, (4) percentage of single-parent households with children under 18.

\item Ranking: minority status and language - Computed following the same procedure as `ranking: socioeconomic status' except the features used were: (1) percentage of people who are not white, non-Hispanic, and (2) percentage of people, age 5 and older, who self-reported speaking English `less than well'.

\item Ranking: housing and transportation - Computed following the same procedure as `ranking: socioeconomic status' except the features used were: (1) percentage of housing in structures with more than 10 units, (2) percentage of mobile homes, (3) percentage of occupied housing with more people than rooms, (4) percentage of households with no vehicle available, and (5) percentage of people in institutionalized group quarters.

\item Index of relative rurality - measure of the level of rurality of a county based on population size, density, remoteness, and built-up area, where 0 indicates that a county has a low level of rurality, i.e. urban, and 1 indicates that a county is the most rural.

\item ICU beds - number of ICU beds available in each county,

\item Nursing home population - the nursing home population in each county. 
\end{enumerate}

Next, we have COVID-19 specific data, including  
\begin{enumerate} 

\item Number of testing locations - testing locations available in each county,

\item Google mobility score - score representing the amount movement (i.e. daily travel) was decreased during state or county shutdowns,

\item State closure status - the status of state shutdowns, 

\item School closure status - the status of school closures.
\end{enumerate}

This second set of features represents both the state and county's policies with respect to the COVID-19 pandemic through testing locations and state and school closure statuses, and it represents the public's response to the pandemic in each county through the Google mobility score. This mobility score is an aggregate measure of the amount that movement between locations such as homes and grocery stores changed due to the pandemic. A county in which the population severely limited their movements at the beginning of the pandemic will have a high mobility score, while a county which remained active will have a low score.

The final group of features, listed below, represents quantities that were engineered from the time-series of cumulative confirmed cases and deaths in each county, from January 22 to August 8 (the day of data collection). Note that January 22 is around the date of the first confirmed COVID-19 case in the US. In order to make the clustering easier, we took the time-series data and described it with a series of ``summary" features which attempt to capture important aspects of the patterns discernible in the time-series. These features include: 
\begin{enumerate} 
\item Case growth rate to first peak - daily rate of change of case numbers from January 22 to April 12 (around the first peak of COVID-19 cases in the US); this metric represents the growth rate of case numbers in each county early in the pandemic

\item Last month case growth rate - daily rate of change of case numbers from July 8 to August 8; representing the growth rate of case numbers in each county later in the pandemic

\item New cases April 12 - number of new confirmed cases reported in each county on April 12, 

\item Deaths April 12 - number of new confirmed deaths reported in each county on April 12, 

\item New cases July 23 - number of new confirmed cases reported in each county on July 23 (around the second peak of COVID-19 cases in the US), 

\item Deaths July 23 - number of new confirmed cases reported in each county on July 23, 

\item Cumulative cases August 8 - number of total cumulative confirmed cases in each county on August 8 (the date of data collection),  

\item Cumulative deaths August 8 - number of total cumulative deaths in each county on August 8 (the date of data collection). 

\end{enumerate} In addition to the new case/death numbers on April 12 (around the first USA-wide peak) and July 23 (around the second USA-wide peak), we also included growth rates of the new case values early in the pandemic and in the month leading up to August 8, and the total cumulative cases/deaths in each county up to August 8. These features aim to capture how the number of confirmed COVID-19 cases and deaths were changing throughout the summer, and also what the raw values looked like during the US case peaks.

\section{Evaluation Metrics and Interpretation Techniques}
Below we describe various methods that we used both for evaluating the clustering methods that we implemented and for interpreting the results of these methods. The content of this section comes from \cite{scikit} and \cite{tan_steinbach_karpatne_kumar_2019}.

\subsection{Evaluation Metrics}
In order to determine which clustering techniques perform well on our data set, we outline several evaluation metrics. While these evaluation metrics cannot offer a guarantee of accuracy since there is no definitive ground truth, they allow us to determine whether or not the clusters we see might represent meaningfully distinct groups of counties. In addition, these evaluation methods can help tune hyperparameters when necessary (e.g. determining the number of clusters for $K$-Means clustering). Below we list some of the metrics we used. 
\begin{itemize}
\item The \emph{silhouette score} gives the average over all points in our dataset of how similar each datapoint is to its own clusters compared to the next closest cluster. Here, the values range from $-1$, indicating clusters that do not separate the datapoints well, to $1$, indicating clusters that clearly distinguish the datapoints.


\item The \emph{Calinski-Harabasz Index} is defined as the ratio of the sum of intra-cluster dispersion and inter-cluster dispersion for all clusters, where dispersion is a measure of distance. The index is high when clusters are dense and well-separated.

\item The \emph{Davies-Bouldin Index} indicates the average ``similarity" between clusters by comparing the distance between clusters with the size of the clusters. Here, 0 is the smallest possible score, and values closer to 0 represent better clusterings.

\item The \emph{elbow method}, also called \emph{distortion}, is a heuristic used to help determine the number of clusters to use for a particular data set. This method consists of plotting the explained variation as a function of the number of clusters, and picking the ``elbow'' of this resulting curve as the number of clusters to use.

\item The \emph{Bayesian information criterion (BIC)} is a criterion for model selection among a finite set of models, specifically for Gaussian mixtures parameter selection. The BIC attempts to choose the model with the highest likelihood while penalizing models with more parameters. The model with the lowest BIC is preferred. 

\item The \emph{Akaike information criterion (AIC)} estimates the information lost by a given model. The less information a model loses, the higher the quality of that model. Similar to the BIC, the AIC tries to find the model with the highest likelihood while penalizing for adding more parameters to the model. However, the penalty for adding more parameters is lower for the AIC than for the BIC.
\end{itemize}

\subsection{Interpretation Techniques} 
Below, we list a few methods we used to interpret the results of applying a particular clustering method. While these methods will not necessarily tell us about how the clusters were created, they can help us understand what makes counties in one cluster similar and what makes two clusters different.

\begin{itemize}
    \item We can find which features have values that differ the most between cluster. That is, if the average values of a feature differ significantly between clusters, that variable can help us distinguish those clusters.
    
    \item We can extract a feature importance score from building a random forest, and then select the features with the highest scores. For this approach, we set the cluster labels as the target value and build a random forest. Then, based on the resulting random forest, we can extract an importance score for each feature. We can interpret the features with the highest scores and features that can help distinguish our clusters.
    
    \item We can describe our clusters or groups of clusters using a decision tree. Here, we build a decision tree using the cluster labels as the output variable. Then, we can see which clusters were grouped near by based on the decision tree, and use the feature value splits in the tree to interpret the clusters or groups of clusters.
    
    \item To see if our clusters represent natural breaks in particular features, we can use the Jenks Optimization Method to cluster using each feature individually. Jenks Optimization Method does single variable clustering. This method uses an iterative approach to group data by gradually minimizing variance within classes and maximizing variance between classes, which is similar to $K$-Means clustering on multiple features. Then, we can use the $v$ measure score to see how similar the clustering from the Jenks Optimization method is to our original clustering, generated by using all features. A higher $v$ measure score indicates a higher agreement of two clustering approaches, which would suggest that the clusters we found originally represent natural breaks in each feature used for the Jenks Optimization method.
        
\end{itemize}

\section{Clustering Techniques}
In this section, we describe the various clustering methods that we implemented in our work. Since one motivation for this project was to learn about different clustering techniques, we tried a wide range of approaches. These include $K$-means and other prototype based methods (Sections 4.1, 4.2, 4.3, 4.4), hierarchical clustering (Section 4.5), and density-based methods (Section 4.6). The content of these sections comes from \cite{scikit} and \cite{tan_steinbach_karpatne_kumar_2019}. We also use \cite{km++} for the $K$-means and other prototype-based approaches, and we use \cite{optics}, \cite{charlon}, and \cite{dbscan} for the section on OPTICS.

\subsection{K-Means Clustering}
The $K$-means algorithm clusters data by separating samples into $K$ groups of equal variance. The algorithm divides a set of samples into disjoint clusters, each described by the mean of the samples in the cluster, called the ``centroid." Note that the centroid may not be an actual sample in the cluster. The $K$-means algorithm has the following four steps: 
\begin{enumerate}
\item Randomly initialize $K$ samples from the dataset as centroids.
\item Assign the rest of the samples in the dataset to their nearest centroid.
\item Update the centroids by taking the mean value of all the samples in each cluster.
\item Repeat steps $2$ and $3$ until the clusters stop changing. 
\end{enumerate}
For this algorithm, $K$ must be specified before starting this algorithm. In the results section below, we describe the method we used to select a appropriate value for $K$.

We note that $K$-means performs poorly on elongated clusters because within-cluster sum-of-squares is small only when clusters are grouped densely. In addition, within-cluster sum-of-squares can easily be large in high-dimensional spaces since datapoints are often more spread out. In order to address this latter issue, we can perform Principle Component Analysis (PCA) in order to reduce the dimensions of the sample space prior to using $K$-means clustering.

\subsection{Mini Batch K-Means Clustering}
Mini Batch $K$-Means is a variant of the $K$-means algorithm with the same first step but where we update the centroids using only some of the data at a time. This algorithm runs as follows:
\begin{enumerate}
\item Randomly initialize $K$ samples from the dataset as centroids.
\item Draw $b$ samples randomly from the dataset to form a mini-batch, and assign each sample to the nearest centroid.
\item Update the centroid of each cluster by taking the average of the newly added samples from that mini-batch as well as all the previous samples assigned to that centroid.
\item Repeat steps 2 and 3 until the clusters stop changing.
\end{enumerate} The results of Mini Batch $K$-Means are often similar to those of $K$-means, but the mini batch version of the algorithm converges faster.

\subsection{Fuzzy C-Means Clustering}
Fuzzy clustering is a form of clustering in which each data point can belong to more than one cluster. Fuzzy $c$-means is one of the most common such methods. It is similar to the $K$-means algorithm, but it also involves a coefficient value for each pair of a datapoint and a cluster, representing the degree to which that datapoint falls into that cluster. Fuzzy $c$-means uses the following steps:
\begin{enumerate}
\item Choose a number of clusters $c$.
\item Randomly initialize coefficients for each sample for being in each cluster.
\item Compute the centroid for each cluster by averaging over all datapoints, where we weight each datapoint by its coefficient.
\item Compute the new coefficients for each datapoint-cluster pair. 
\item Repeat steps $3$ and $4$ until the algorithm converges. 
\end{enumerate} 

\subsection{Gaussian Mixture Models}
A Gaussian mixture model is a probabilistic model that assumes our datapoints are generated from a mixture of a finite number, say $k$, of Gaussian distributions with unknown parameters. Each Gaussian distribution in the mixture is composed of the following parameters:
\begin{itemize} 
\item A mean $\mu_k$ defining the center of the distribution, 
\item A covariance matrix $\Sigma_k$ defining the variance of the distribution, and
\item A mixing probability $\pi_k$ defining how big the Gaussian function will be, or the relative sizes of the different Gaussians.
\end{itemize} 
We can then find these parameters by iteratively updating estimations for them. In Gaussian mixture model selection process, we also need to specify the covariance type and the number of Gaussians to use. We can use BIC and AIC to help us select both of these parameters. 

\subsection{Hierarchical Clustering}
One common method for finding structure in unlabeled data is hierarchical clustering. Unlike many other clustering approaches, this technique does not find one specific number of clusters within a dataset, but instead finds all numbers of clusters at once. Then, from the resulting hierarchy, we can select the number of clusters that we want to use.

There are two main types of hierarchical clustering: agglomerative and divisive. Agglomerative clustering starts with all observations in their own cluster, and sequentially joins the two closest clusters until every observation is in one cluster together. Divisive clustering, on the other hand, starts with a single cluster of all the observations and sequentially splits off one observation. Here, we focus on the agglomerative approach.

When applying agglomerative clustering, we need to specify (1) a distance metric between observations and (2) a definition of distance between clusters, rather than just individuals. In our work on this project, we used \texttt{linkage} from \texttt{scipy.cluster.hierarchy}, which has a number of available distance metrics, and cluster distances, also called linkages. For distance metric, we tested almost all that are available in \texttt{pdist} from \texttt{scipy.spatial.distance}, including Euclidean, Minkowski, cityblock (or Manhattan), cosine, and squared Euclidean distances. For distance between clusters, we tested all available options, including Ward’s, complete, average, and single linkages. We also specified a final desired number of clusters, so that as we are iteratively merging clusters, we stop once we have this number remaining.

\subsection{OPTICS Clustering}

In this section, we will describe the OPTICS clustering algorithm. First, though, we will introduce the DBSCAN clustering method.

\subsubsection{DBSCAN}
Density-based spatial clustering of applications with noise (DBSCAN) is a density-based non-parametric algorithm which locates regions of high density in the data that are separated from one another by regions of low-density. There are several approaches to defining the density of the data points. In this section, we will mainly use the center-based approach for defining the density.

In the center-based approach, we have \texttt{MinPts} and \texttt{Eps}, both of which are user-specified parameters representing the number of points and the radius. Then, we classify all the data points in the given space into one of the following three categories:

\begin{itemize} 
\item \textbf{Core Points:} We label a datapoint as a core point if there are at least \texttt{MinPts} datapoints (including the point itself) within a distance of radius \texttt{Eps} of this point.

\item \textbf{Border Points:} we label a datapoint as a border point if it is in the neighborhood of a core point. That is, it is within \texttt{Eps} of a core points, but there are fewer than \texttt{MinPts} datapoints within a distance of \texttt{Eps} from this border point.

\item \textbf{Noise:} We label a datapoint as noise if it is neither a core point nor a border point.
\end{itemize}

After labeling each point, we then form our clusters. We group two core points into one cluster if the distance between them is less than \texttt{Eps}. We group a border point into a cluster if the border point is within a distance of \texttt{Eps} from a core point in that cluster. Finally, we group all the noise points together into one cluster. The resulting clusters are the output of the DBSCAN algorithm.

\subsubsection{OPTICS}\label{sec:optics1}
We notice that DBSCAN uses a fixed distance threshold (\texttt{Eps}) which can not detect clusters of varying densities, which can sometimes lead to poor clustering. Ordering Points To Identify the Clustering Structure (OPTICS) has a similar algorithm to DBSCAN. However, OPTICS addresses this problem of a fixed distance by using a steepness threshold for \texttt{Eps}.

In the OPTICS algorithm as in DBSCAN, we specify a value \texttt{MinPts}, which gives us a lower bound on the number points in the neighborhood of some point in order for us to consider that point, and a value \texttt{Eps}, which is the maximum radius of a neighborhood around a point that we will consider. Then in the OPTICS algorithm, for each datapoint $p$, we determine its core distance, which is the minimum radius size so that there are \texttt{MinPts} points within that distance. Note that if this radius is larger than \texttt{Eps}, we set the core distance of $p$ to be undefined. Then for any point $q$ within the \texttt{Eps}-neighborhood of $p$, we set the reachability distance from $p$ to $q$ to be the maximum of the core distance of $p$ and of the distance from $p$ to $q$. If $q$ is further than \texttt{Eps} distance from $p$, we let the reachability distance from $p$ to $q$ be undefined.

\begin{figure}
 \centering
 \includegraphics[width=0.5\linewidth]{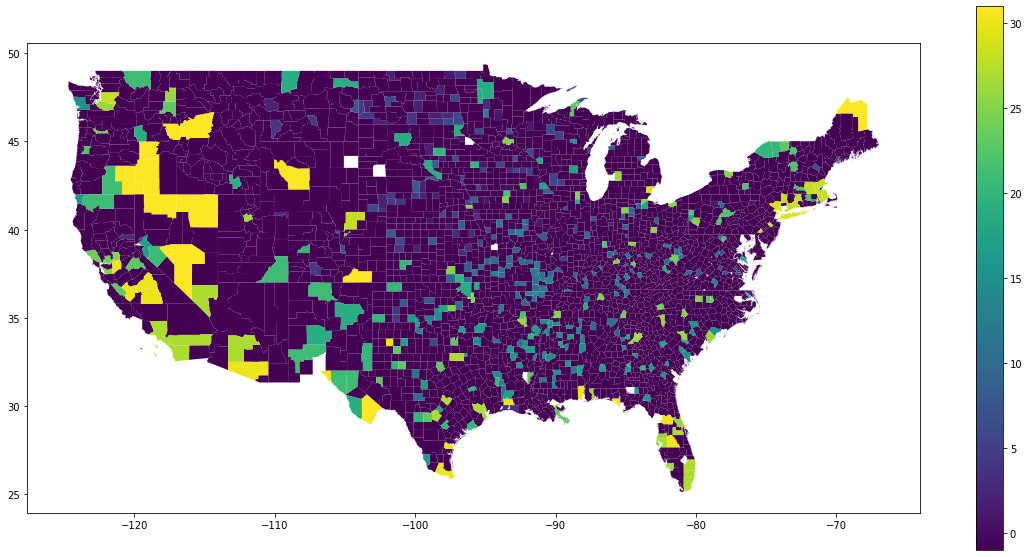}
 \captionof{figure}{OPTICS Reachability Graph Example \cite{optics}}
 \label{fig:optics}
\end{figure}

Next, we order our datapoints so that points near each other with respect to reachability distance are near each other in the ordering. This produces a reachability graph as in Figure \ref{fig:optics}, where reachability score is plotted against the datapoint ordering. The spikes in this graph of high reachability scores correspond to less densely connected pointed, and thus indicate potential boundaries between clusters. Then, we can set different reachability thresholds and set the resulting clusters to be all points between spikes of at scores higher than the threshold.

To apply the OPTICS Clustering to our data, we used the package from
\\
\texttt{sklearn.cluster.OPTICS}
which requires two parameters: \texttt{min\_samples} (\texttt{Mps}) and \texttt{metric}. To select these parameters, we implemented a grid search over a range of 2-30 for \texttt{min\_samples} and over almost all the metrics in \texttt{scikit-learn} and \texttt{scipy.spatial.distance}, and selected the best pair of parameters using the silhouette score and Caliski-Harabasz index.

\section{Results}
In the following sections, we describe our results from implementing the clustering algorithms described in Section 4 to the data we gathered described in Section 2.

\subsection{K-means results}
Since the $K$-means clustering method does not work well with high-dimensional data, as mentioned in Section 4.1, we first used principle component analysis (PCA) to visualize the data and reduce dimensionality before clustering. This process showed that by using 8 principle components, we were able to capture approximately 95\% of the variance in our data. This preprocessing step reduced our feature space by more than half.

\begin{figure}
     \centering
     \begin{subfigure}[b]{0.3\textwidth}
         \centering
         \includegraphics[width=\textwidth]{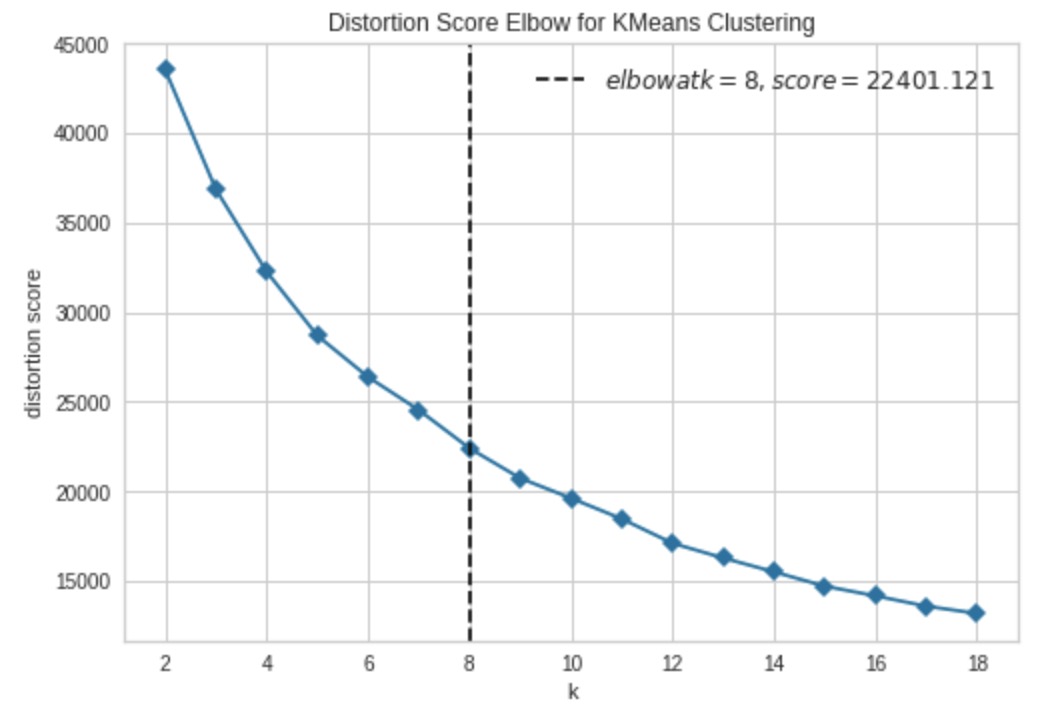}
         \caption{Elbow Score}
         \label{fig:ke}
     \end{subfigure}
     \hfill
     \begin{subfigure}[b]{0.3\textwidth}
         \centering
         \includegraphics[width=\textwidth]{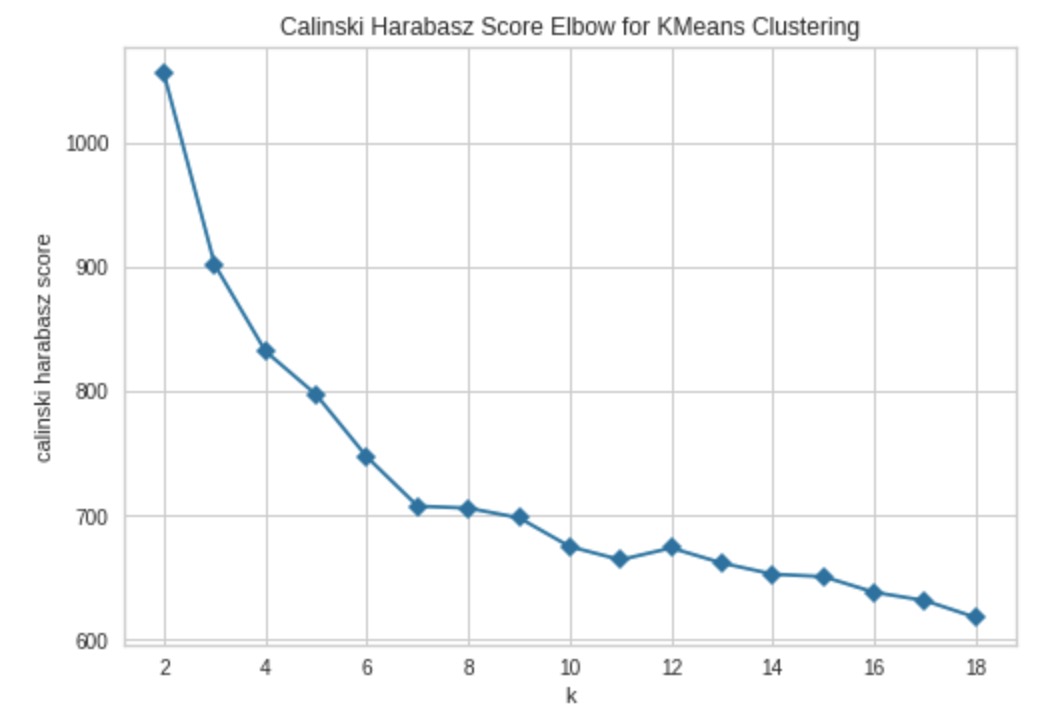}
         \caption{Calinski Harabasz Score}
         \label{fig:kch}
     \end{subfigure}
     \hfill
     \begin{subfigure}[b]{0.3\textwidth}
         \centering
         \includegraphics[width=\textwidth]{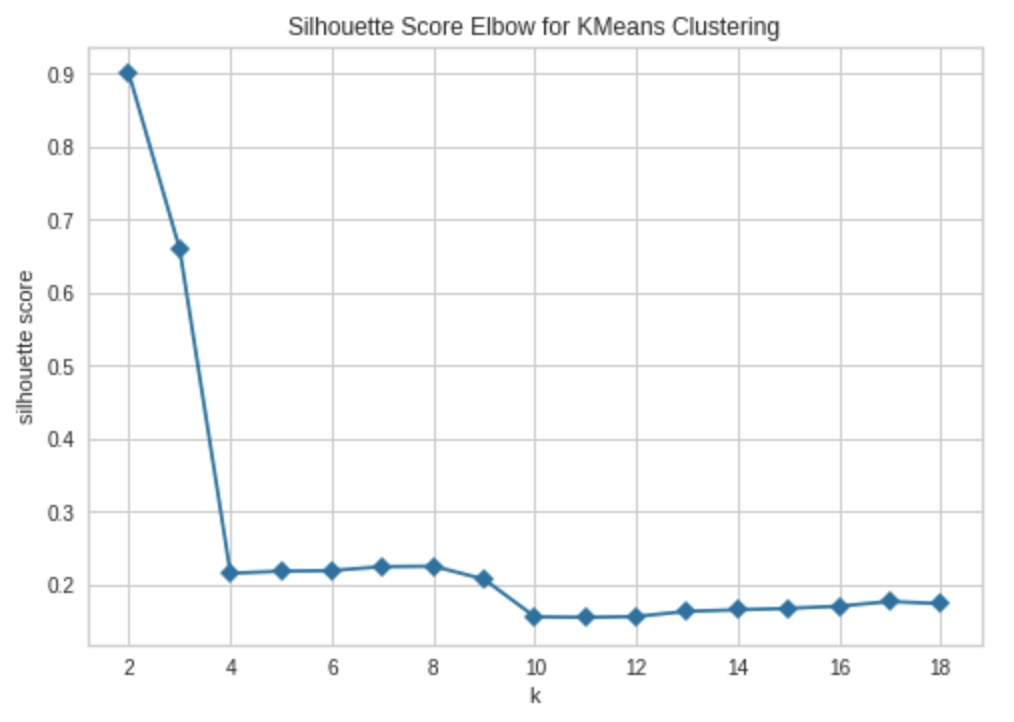}
         \caption{Silhouette Score}
         \label{fig:ksh}
     \end{subfigure}
        \caption{K-Means Evaluation Scores}
        \label{fig:kev}
\end{figure}

Next, we chose $K$, the number of clusters, using the elbow score, silhouette score, and Calinski-Harabasz index. We used the package \texttt{yellowbrick} to calculate and visualize these scores since it has an indicator that automatically shows the optimal number of components based on each evaluation methods that balances a high score with using more clusters. The optimal number of components according to the elbow score is 8. The Calinski-Harabasz index has a similar shape to the elbow score plot, which decreases quickly between 4 and 7 clusters, and then decreases more slowly afterwards. This suggests that perhaps around 7 or 8 clusters is reasonable, as well. However, the silhouette score drops off sharply between 2 and 4 clusters and then stays around the same score through at least 18 clusters. We can see the relevant plots of these scores with the corresponding best number of features indicated in Figure \ref{fig:kev}.

\begin{figure}
 \centering
 \includegraphics[width=\linewidth]{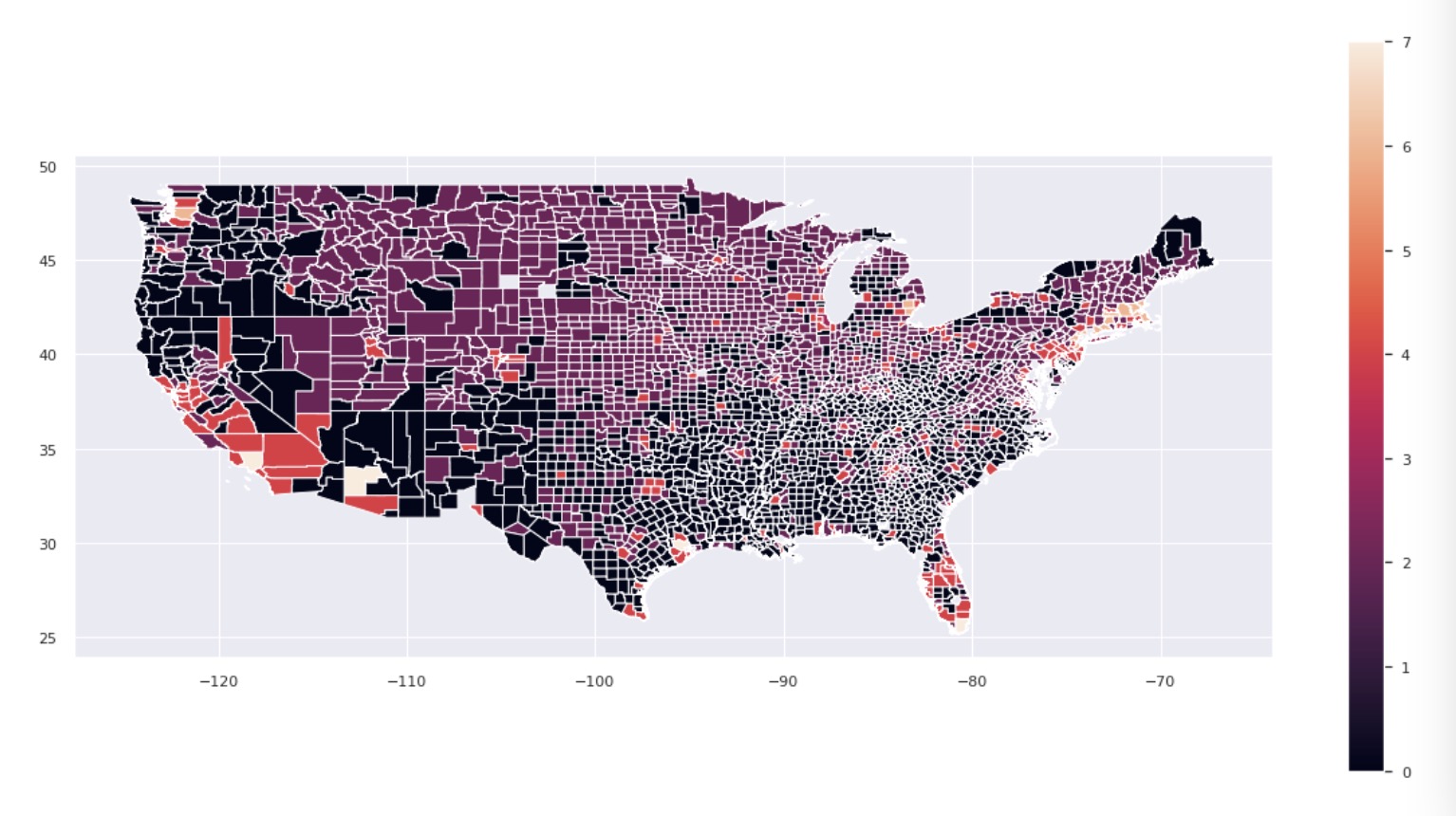}
 \captionof{figure}{K-Means Map}
 \label{fig:kmap}
\end{figure}

Because both elbow score and Calinski-Harabasz index both suggested that around 8 clusters was a good choice, we choose this the number of clusters to use. After performing $K$-means clustering with 8 clusters, we can visualize the resulting clusters in the map in Figure \ref{fig:kmap}, where the color of a county corresponds to the cluster that it is in. We note that many counties in the southern half of the United States, as well as those in the west appear to be clustered together, and many counties in the Midwest as well as the Northeast appear to be clustered together.

To interpret the clusters, first, we selected the important features using random forest feature selection as described in Section 3.2. The top seven important features are listed here:

\begin{itemize}
\item county ranking with respect to socioeconomic status,
\item county ranking with respect to household decomposition and disability,
\item county ranking with respect to minority status and language,
\item county population,
\item cumulative cases as of August 8,
\item number of ICU beds.  
\end{itemize}

\begin{figure}
\centering
\includegraphics[width=0.5\linewidth]{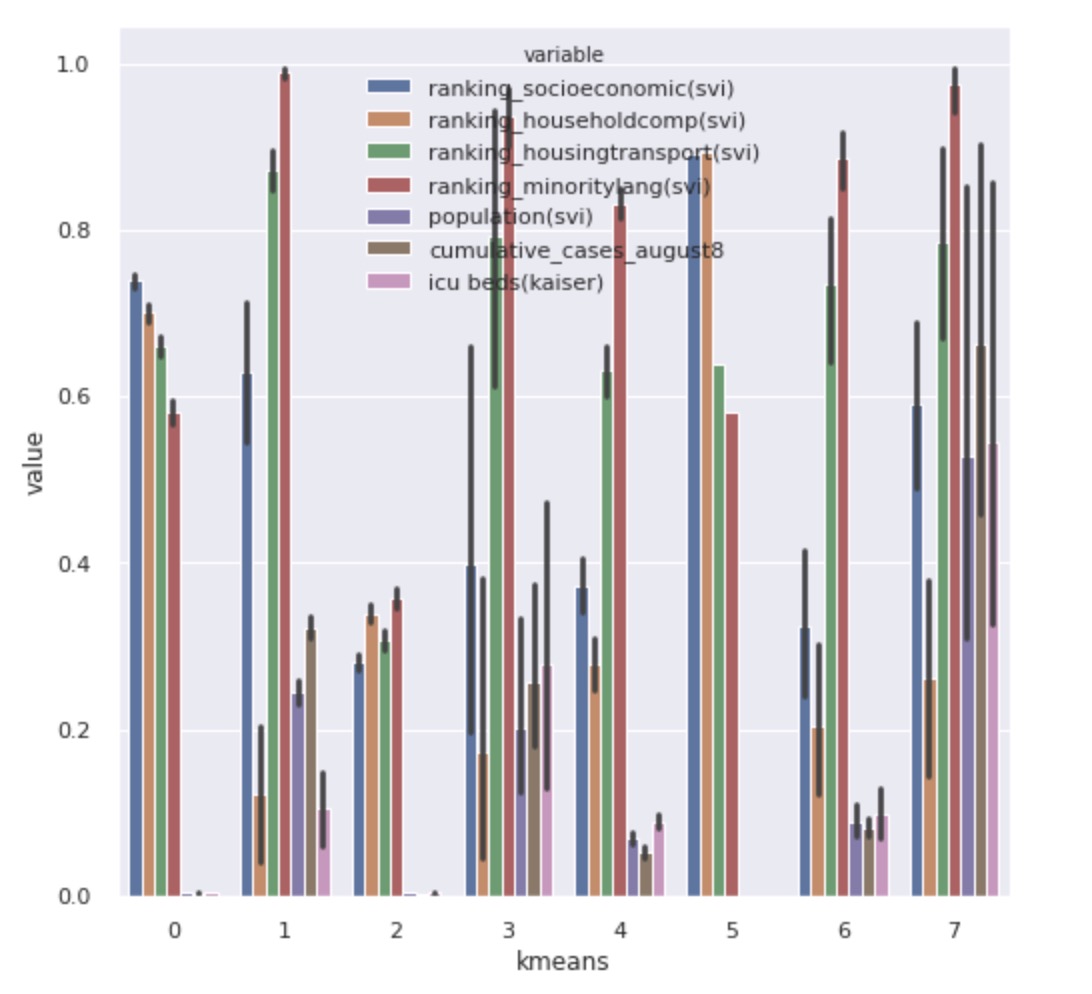}
\captionof{figure}{K-Means clusters Important Features}
\label{fig:kif}
\end{figure}

Figure \ref{fig:kif} shows the average variation of each features among different clusters.

We also used Jenks Optimization Method to cluster each features individually, and then we used the $v$-measure score between each feature clusters and the overall clusters to test the homogeneity between these sets of clusters. A high score indicates a high homogeneity. The features with the highest scores for $K$-Means clustering are: cumulative cases as of August 8, county ranking with respect to socioeconomic status, and county population. This suggests that natural breaks in the values for these features match up with to the results of $K$-means clustering, meaning that our clusters seem to correspond to different values of these features.

\subsection{Fuzzy-C means}
\begin{figure}
     \centering
     \begin{subfigure}[b]{0.3\textwidth}
         \centering
         \includegraphics[width=\textwidth]{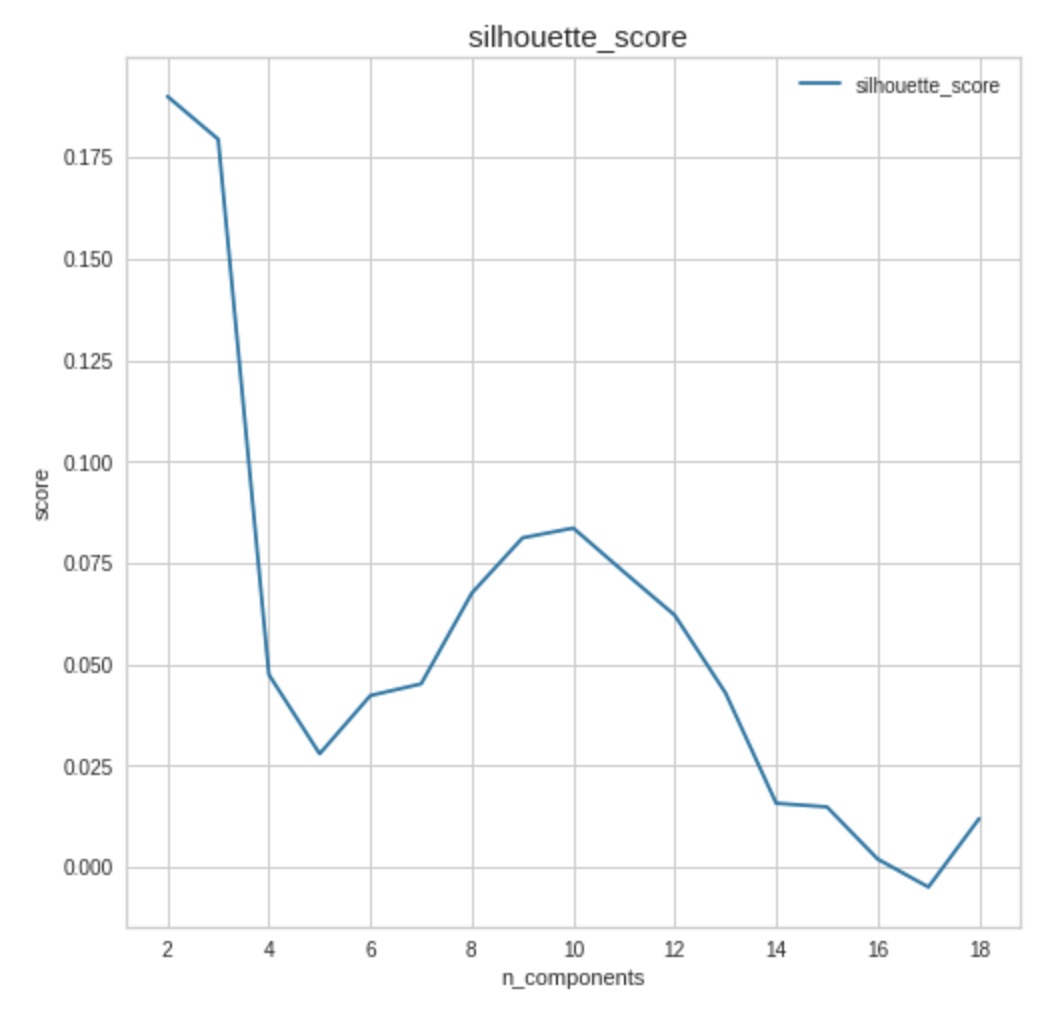}
         \caption{Silhouette Score}
         \label{fig:fcmsh}
     \end{subfigure}
     \hfill
     \begin{subfigure}[b]{0.3\textwidth}
         \centering
         \includegraphics[width=\textwidth]{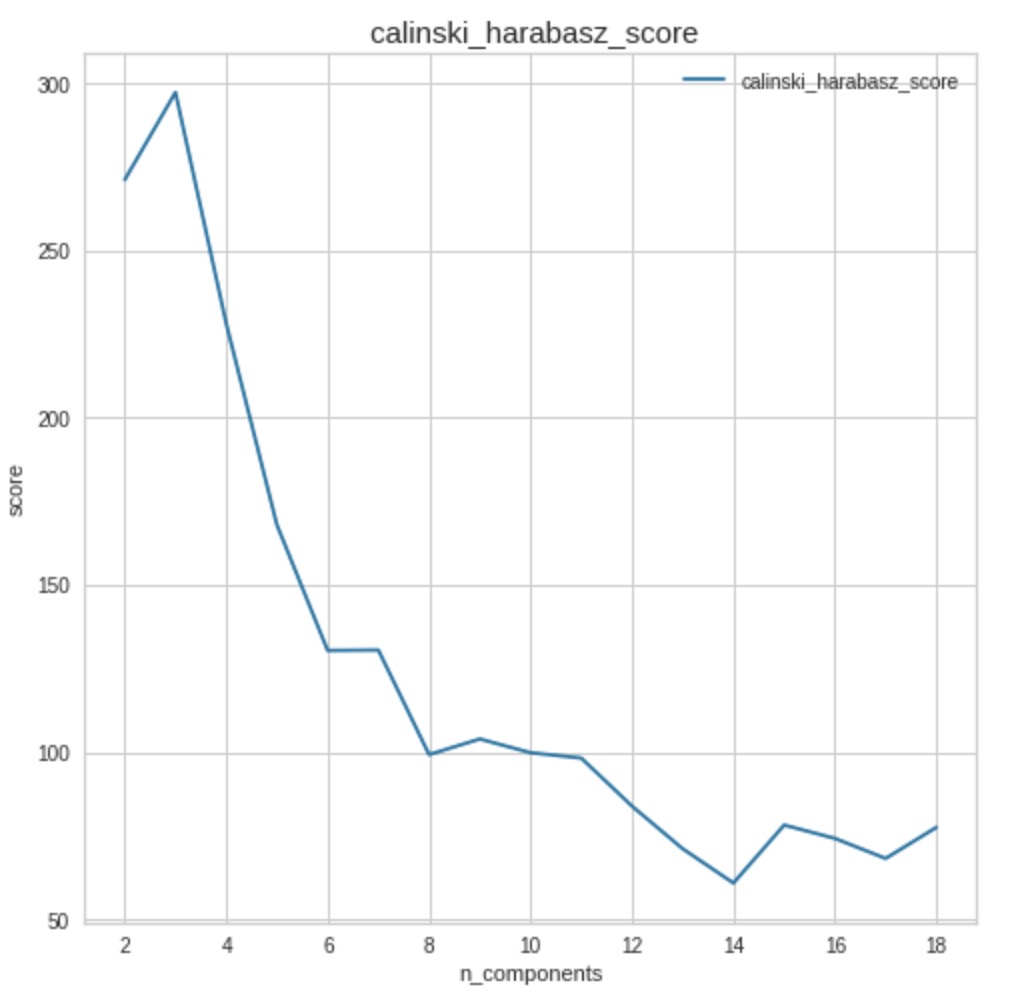}
         \caption{Calinski Harabasz Score}
         \label{fig:fcmch}
     \end{subfigure}
     \hfill
     \begin{subfigure}[b]{0.3\textwidth}
         \centering
         \includegraphics[width=\textwidth]{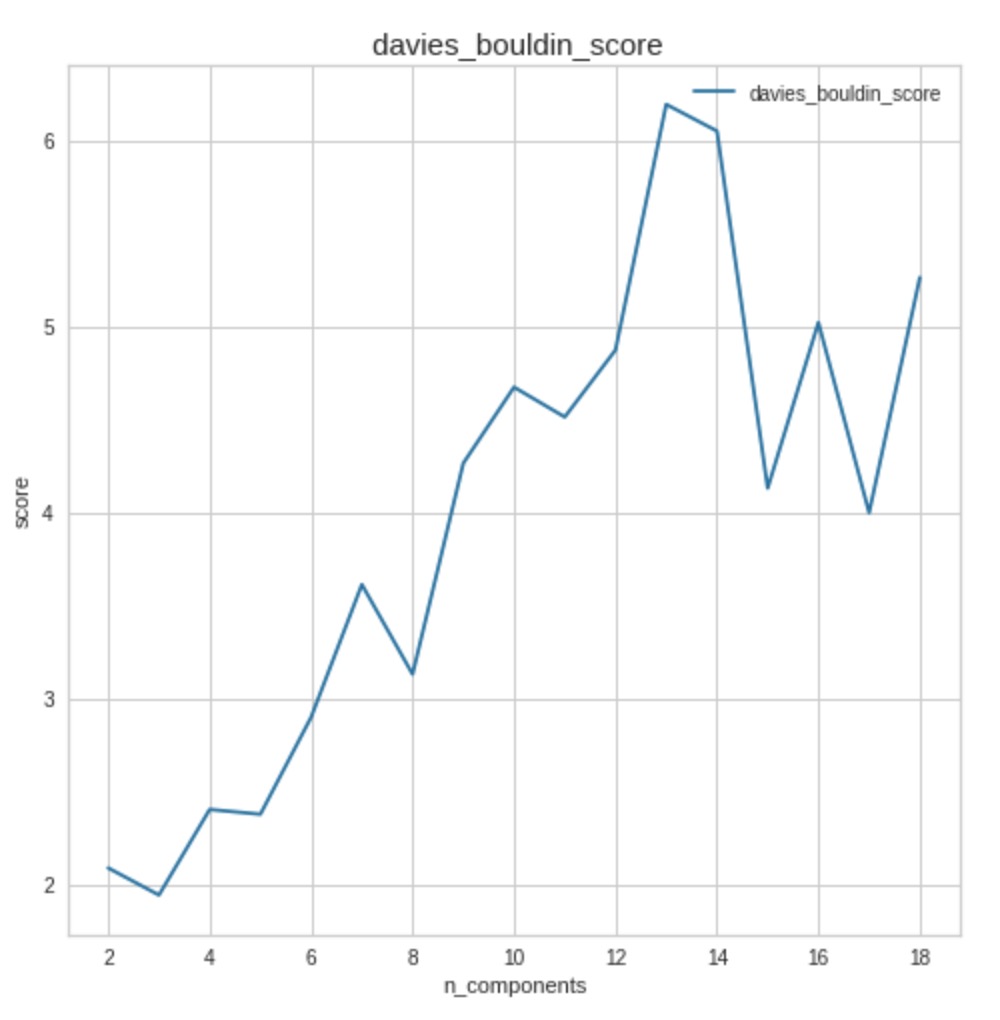}
         \caption{Davies Bouldin Score}
         \label{fig:fcmdb}
     \end{subfigure}
        \caption{Fuzzy-C Means Evaluation Scores}
        \label{fig:fev}
\end{figure}

To select the optimal number of clusters for Fuzzy-$c$ means clustering, we used the silhouette score, the Calinski-Harabasz score, and the Davies-Bouldin score similarly to in the section above. As we can see in Figure \ref{fig:fev}, each of these returned $3$ as the optimal number of clusters.

\begin{figure}
 \centering
 \includegraphics[width=\linewidth]{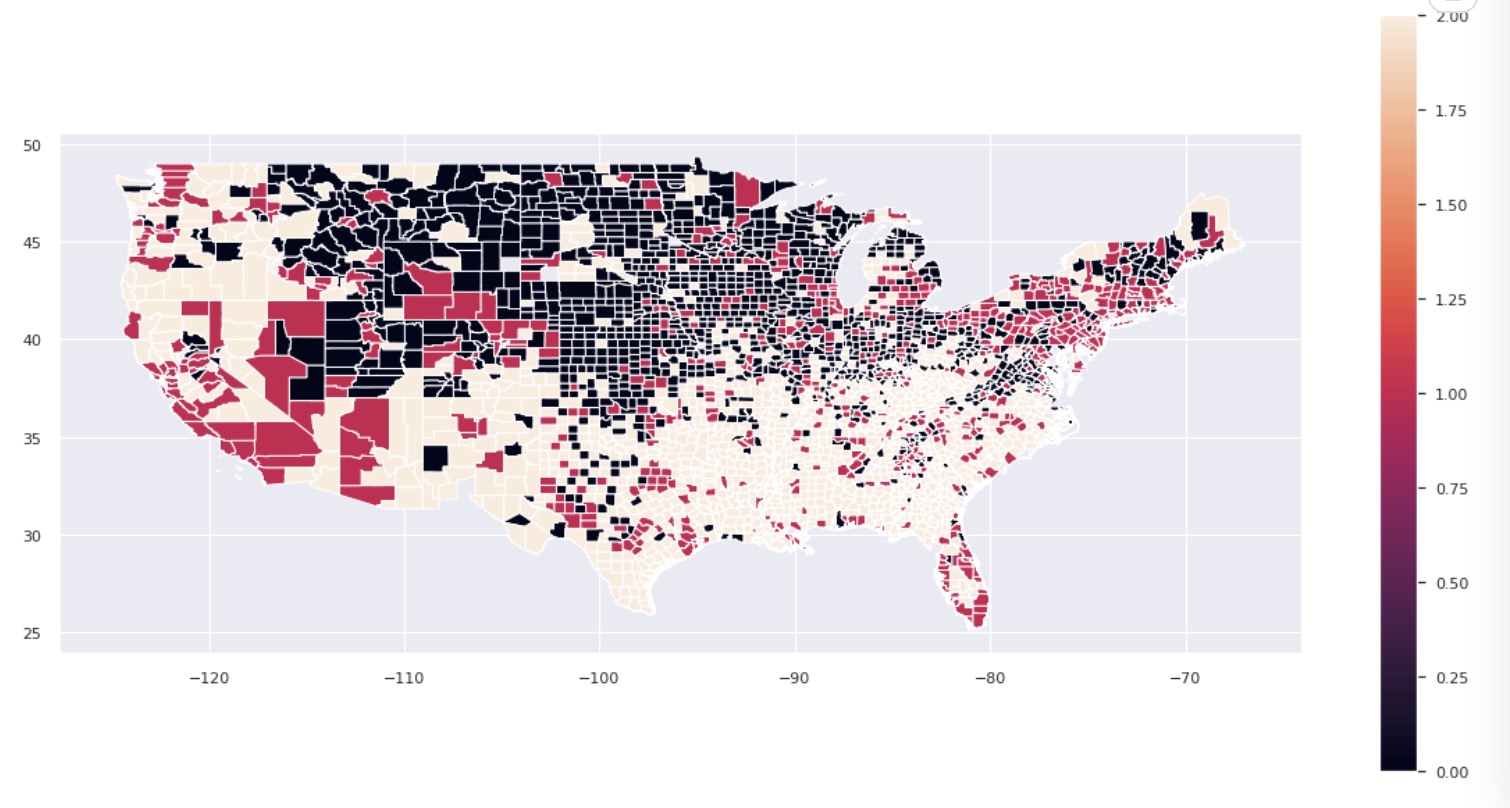}
 \captionof{figure}{Fuzzy-c Means Map}
 \label{fig:examplewithalignment}
\end{figure}

After performing Fuzzy-$c$ means clustering with 3 clusters, we can visualize the resulting clusters in the map in Figure \ref{fig:examplewithalignment}, where the color of a county corresponds to the cluster that it is in. Note that the clusters here appear to be very similar to the clusters we found using $K$-means clustering. This makes sense since these two methods are quite similar.

The top seven important features according to random forest feature selection are listed here: 

\begin{itemize}
\item county ranking with respect to socioeconomic status,
\item county ranking with respect to housing and transportation,
\item county ranking with respect to household decomposition and disability, 
\item county ranking with respect to  minority status and language,
\item county population,
\item index of relative rurality,
\item cumulative cases as of August 8.  
\end{itemize}

\begin{figure}
 \centering
 \includegraphics[width=0.5\linewidth]{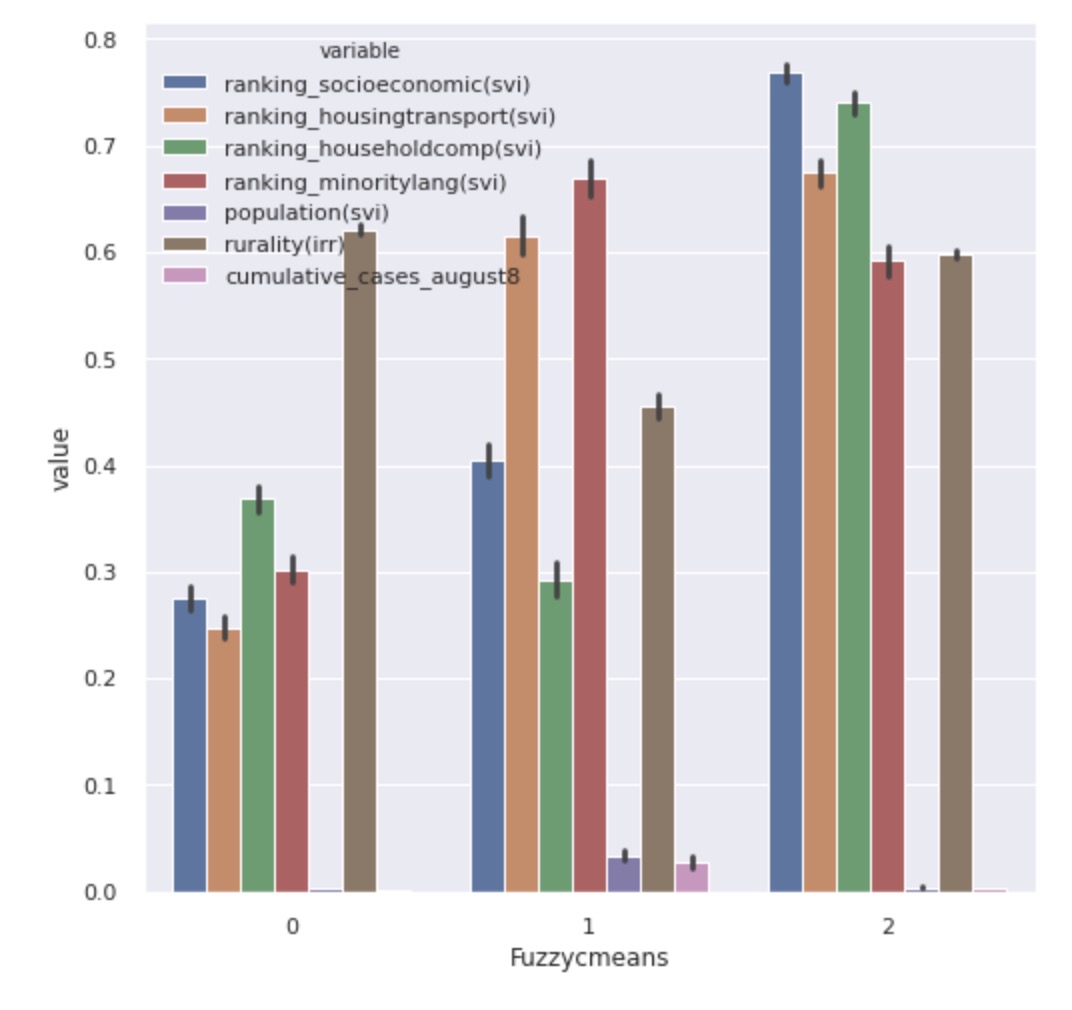}
 \captionof{figure}{Fuzzy-$c$ Means Clusters Important Features}
 \label{fig:examplewithalignment2}
 \end{figure}

Figure \ref{fig:examplewithalignment2} shows the average variation of each features among different clusters. The results show that the values for ranking with respect to socioeconomic status, ranking with respect to housing and transportation, and ranking with respect to household decomposition differ between the clusters. Therefore, we are able to use them distinguish the clusters we got from Fuzzy-$c$ Means clustering.

\subsection{Gaussian Mixture Model}
\begin{figure}
     \centering
     \begin{subfigure}[b]{0.3\textwidth}
         \centering
         \includegraphics[width=\textwidth]{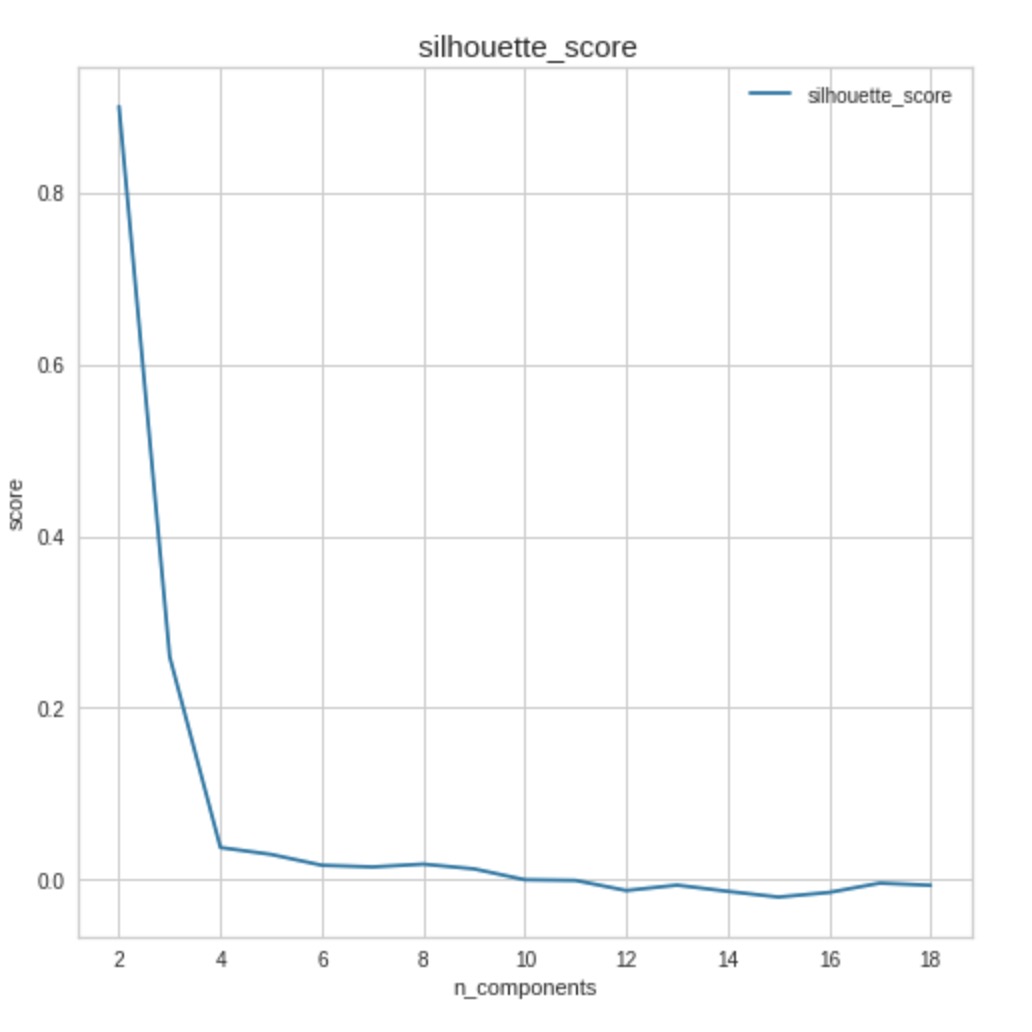}
         \caption{Silhouette Score}
         \label{fig:gmsh}
     \end{subfigure}
     \hfill
     \begin{subfigure}[b]{0.3\textwidth}
         \centering
         \includegraphics[width=\textwidth]{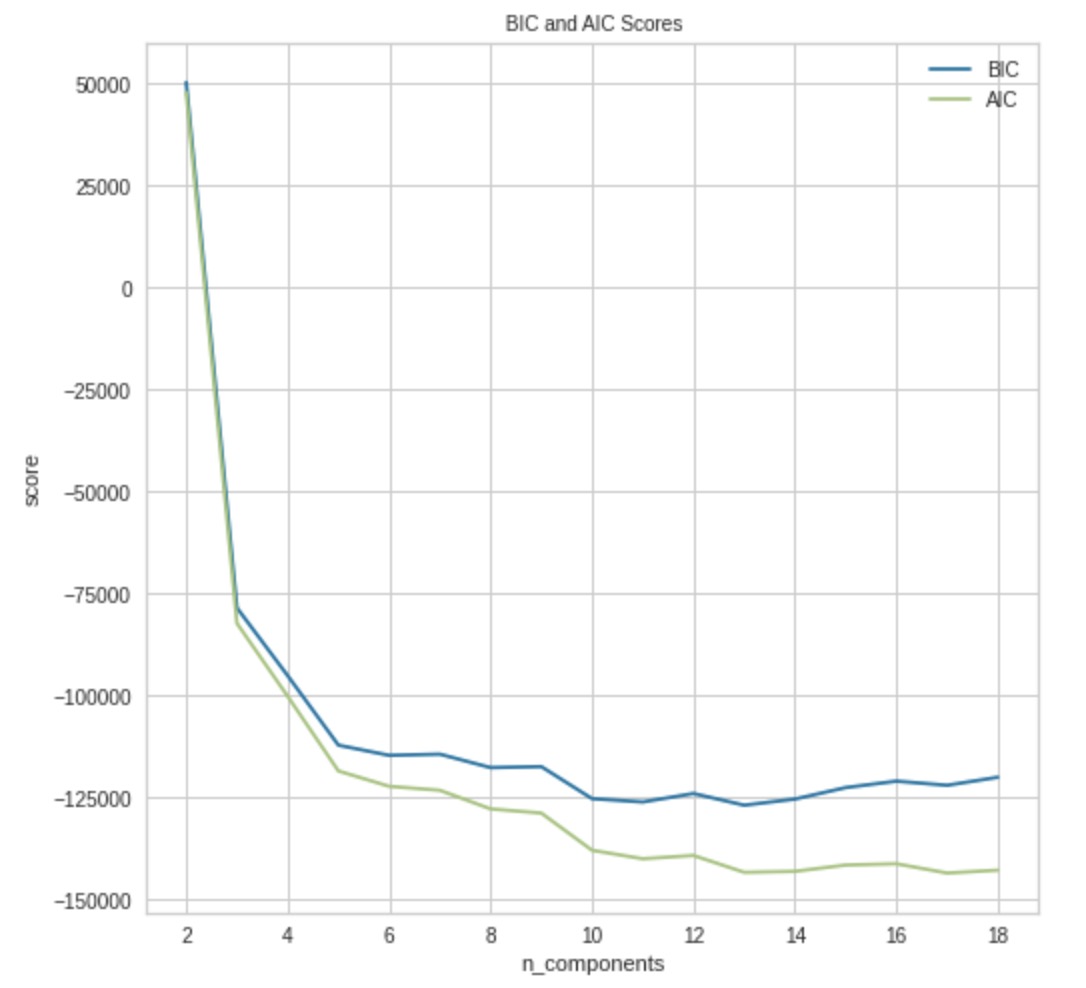}
         \caption{BIC and AIC Score}
         \label{fig:gmbicaic}
     \end{subfigure}
     \hfill
     \begin{subfigure}[b]{0.3\textwidth}
         \centering
         \includegraphics[width=\textwidth]{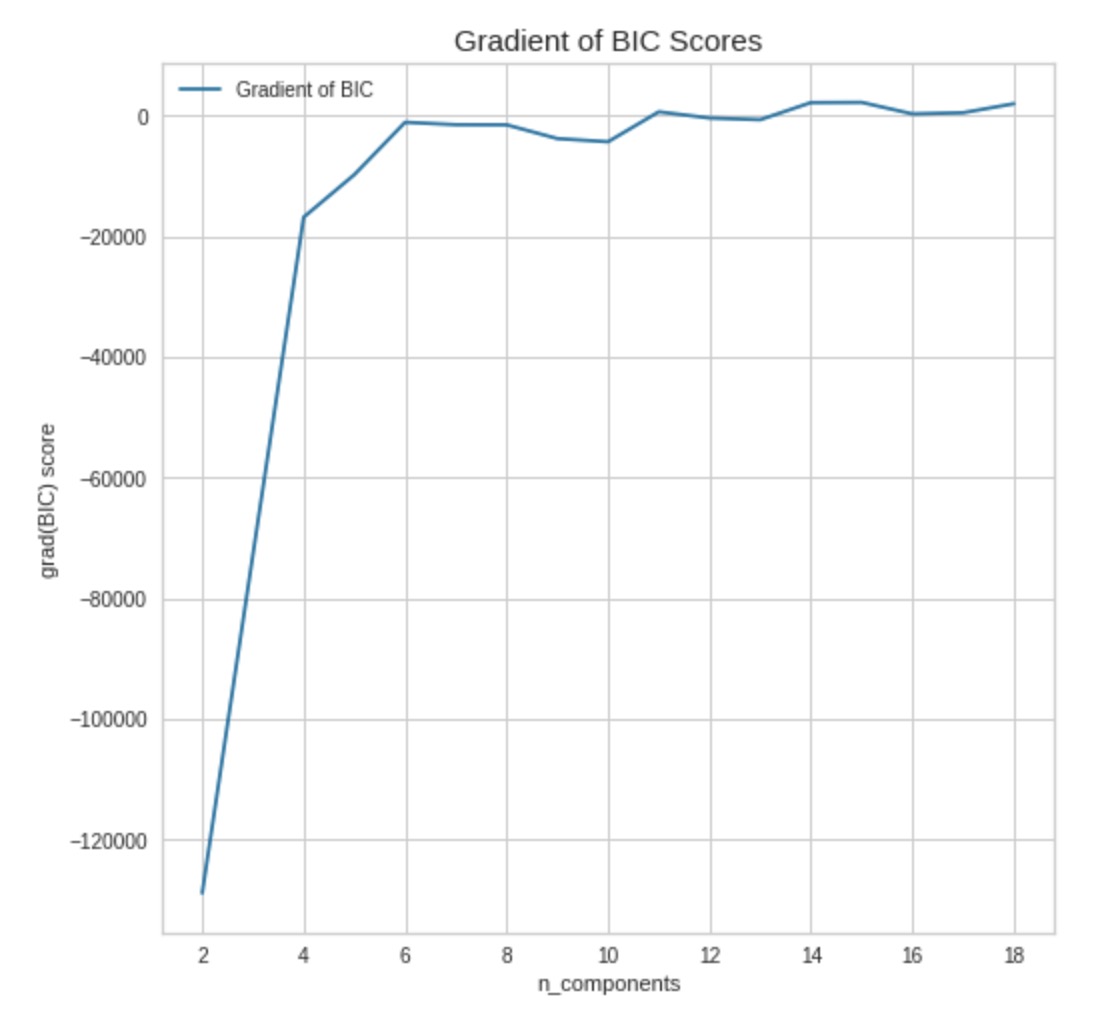}
         \caption{Gradient BIC Score}
         \label{fig:gmgbic}
     \end{subfigure}
        \caption{Gaussian Mixture Model Evaluation Scores}
        \label{fig:gev}
\end{figure}

For the Gaussian Mixture model approach, the evaluation methods we used were silhouette score, BIC, and AIC. BIC and AIC both return an optimal number of clusters of either 3 or 5, but the silhouette score is higher for 3 clusters, meaning the best choice is 3 clusters, as we can see in Figure \ref{fig:gev}.

\begin{figure}
 \centering
 \includegraphics[width=\linewidth]{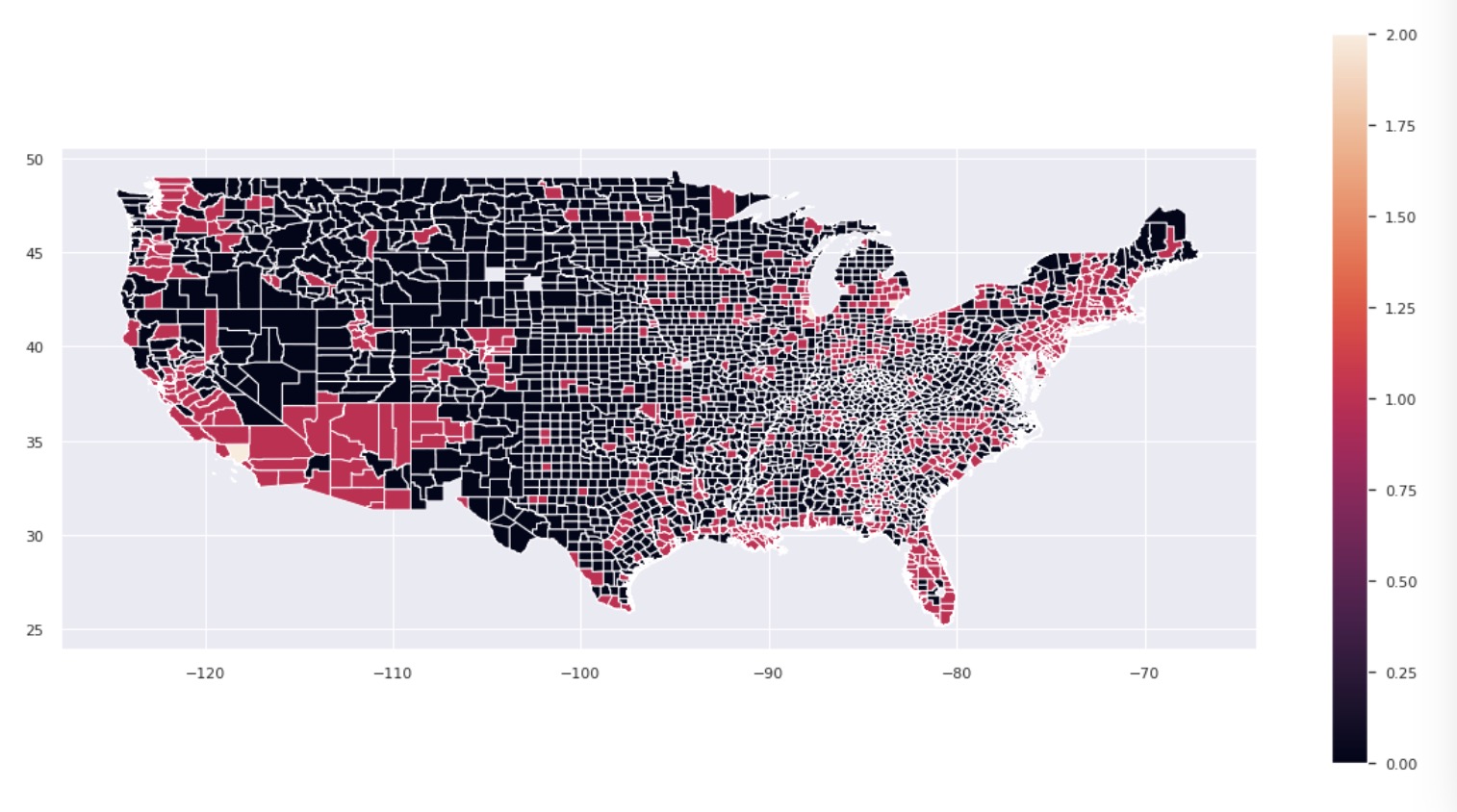}
 \captionof{figure}{Gaussian Mixture Map}
 \label{fig:gmmm}
\end{figure}

After building a Gaussian Mixture model with 3 clusters, we can visualize the resulting clusters in the map in Figure \ref{fig:gmmm}, where the color of a county corresponds to the cluster that it is in. Here we see that almost every county is in cluster 0 or cluster 1, and very few counties are in cluster 2.

The top seven important features according to random forest feature selection are listed here: 

\begin{itemize}
\item new cases on July 23,
\item case growth rate to first peak, 
\item county population, 
\item cumulative cases as of August 8, 
\item cumulative deaths as of August 8,
\item number of ICU beds,
\item index of relative rurality. 
\end{itemize}

\begin{figure}
 \centering
 \includegraphics[width=0.5\linewidth]{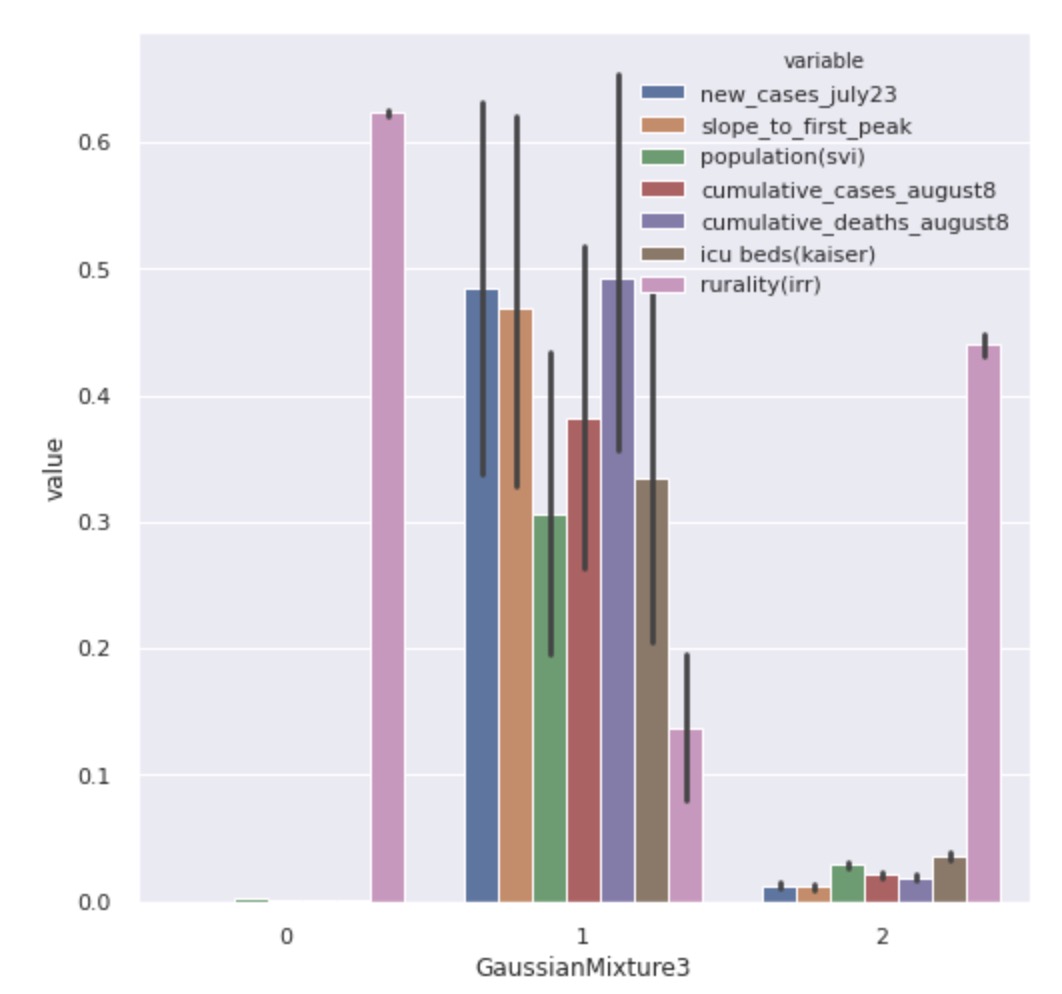}
 \captionof{figure}{Gaussian Mixture Important Features}
 \label{fig:examplewithalignment3}
 \end{figure}

Figure \ref{fig:examplewithalignment3} shows the average variation of each feature across the different clusters. This shows that the average value for index of relative rurality differs greatly between the clusters we got from the Gaussian mixture model. Therefore, we can use this feature to distinguish our resulting clusters, where on average cluster 0 contains the more rural counties, cluster 1 contains the less rural counties, and cluster 2 contains counties somewhere in the middle with regards to rurality.

\subsection{Mini Batch K-Means} 

\begin{figure}
     \centering
     \begin{subfigure}[b]{0.3\textwidth}
         \centering
         \includegraphics[width=\textwidth]{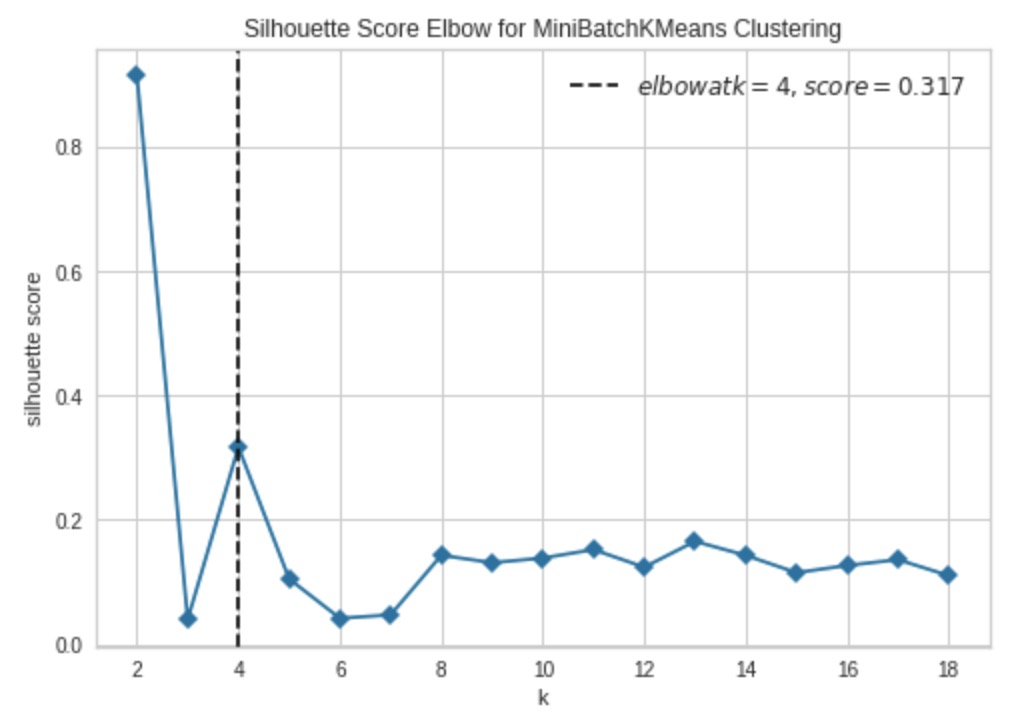}
         \caption{Silhouette Score}
         \label{fig:mbksh}
     \end{subfigure}
     \hfill
     \begin{subfigure}[b]{0.3\textwidth}
         \centering
         \includegraphics[width=\textwidth]{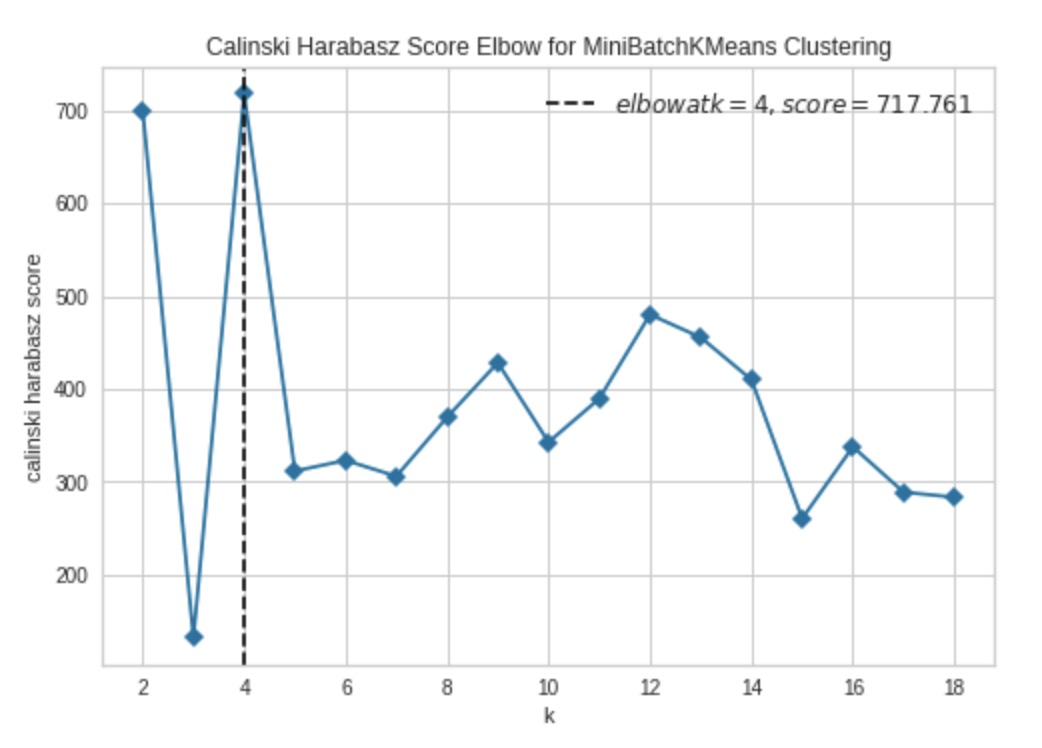}
         \caption{Calinski-Harabasz Score}
         \label{fig:mbkch}
     \end{subfigure}
     \hfill
     \begin{subfigure}[b]{0.3\textwidth}
         \centering
         \includegraphics[width=\textwidth]{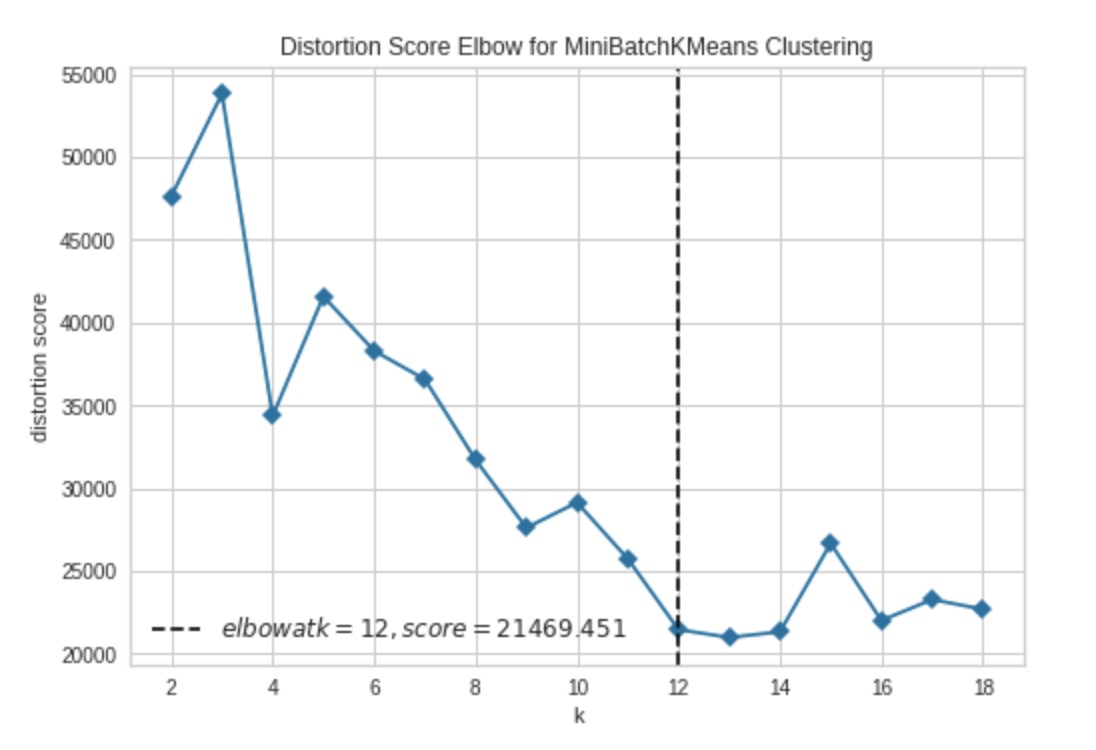}
         \caption{Gradient Elbow Score}
         \label{fig:mbkeb}
     \end{subfigure}
        \caption{Mini Batch K-Means Evaluation Scores}
        \label{fig:mbkev}
\end{figure}

Now, we describe our results form applying Mini Batch $K$-means. In order to determine the number of clusters for this approach, we used the evaluation methods silhouette score, Calinski-Harabasz score, and elbow score. Both silhouette score and Calinski-Harabasz score indicate that 4 is a good choice for the number of clusters, as we can see in Figure \ref{fig:mbkev}. The elbow score indicates that 12 is a good choice for the number of clusters. However, since the corresponding elbow score with 4 clusters is a local minimum, we decided to use 4 clusters for Mini Batch $K$-means.

\begin{figure}
 \centering
 \includegraphics[width=\linewidth]{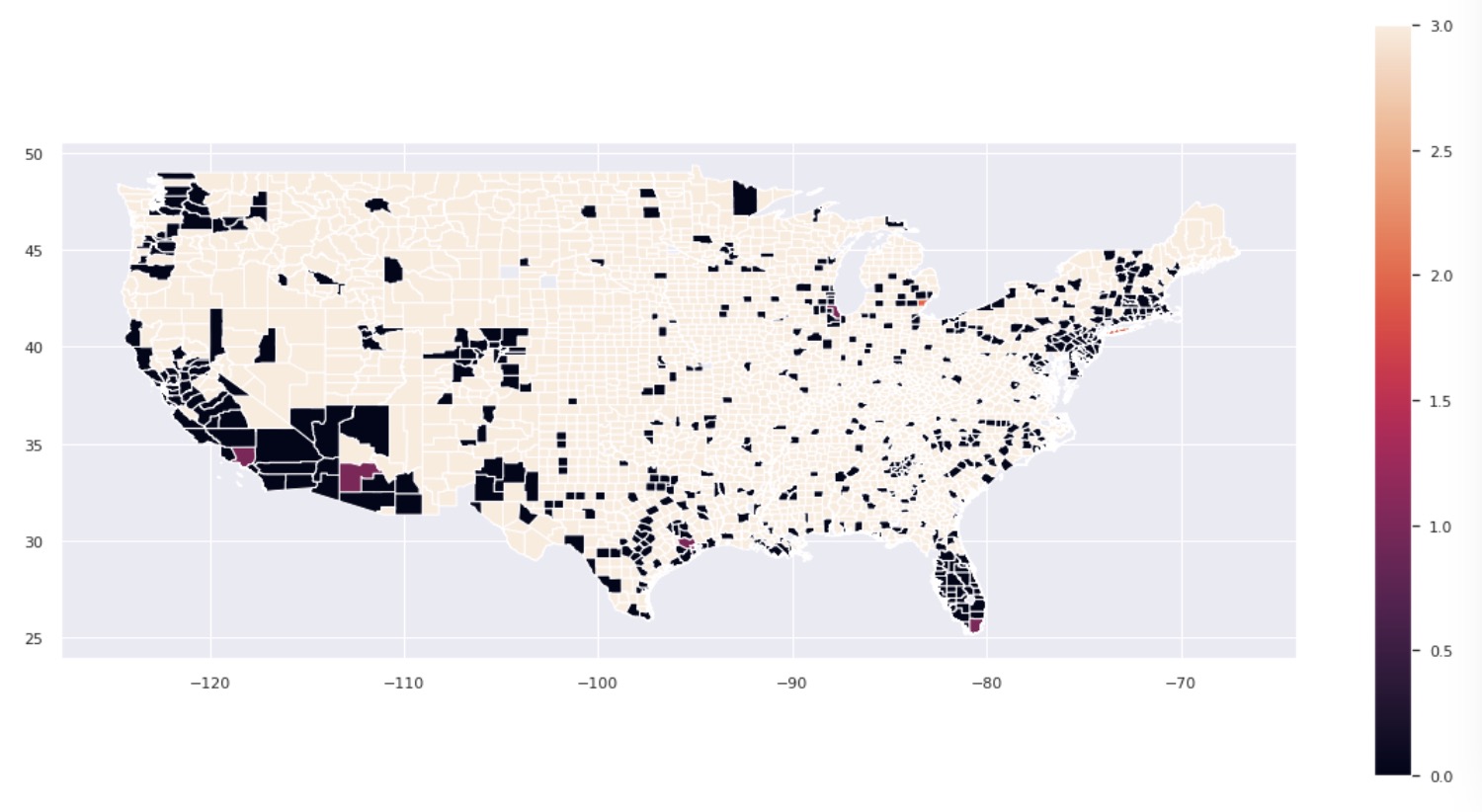}
 \captionof{figure}{Mini Batch $K$-Means Map}
 \label{fig:mbkm}
\end{figure}

After building a Mini Batch $K$-means model with 4 clusters, we visualized the resulting clusters in the map in Figure \ref{fig:mbkm}, where the color of a county corresponds to the cluster that it is in.

The top seven important features according to random forest feature selection are listed here: 

\begin{itemize}
\item county population, 
\item cumulative cases as of August 8,
\item mobility score,
\item index of relative rurality, 
\item number of ICU beds,
\item case growth rate to first peak, 
\item county ranking with respect to minority status and language. 
\end{itemize}

\begin{figure}
\centering
\includegraphics[width=0.5\linewidth]{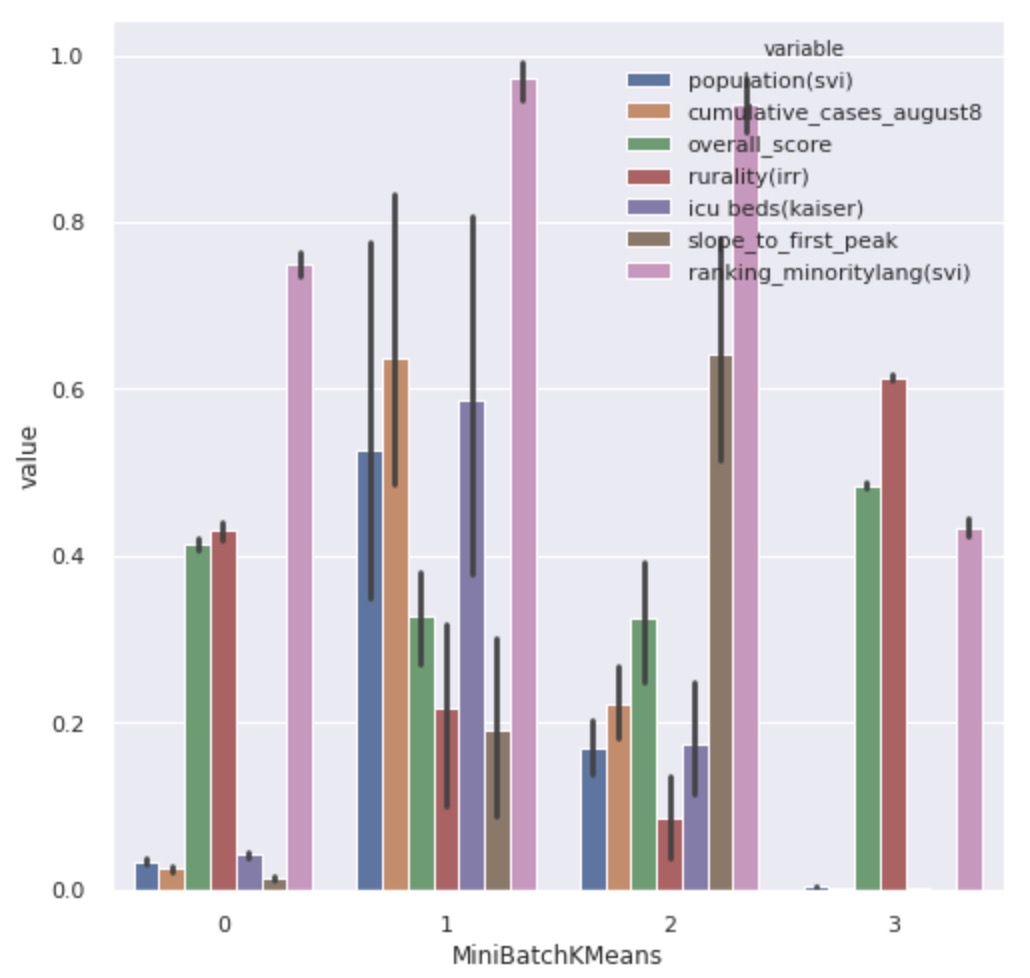}
\captionof{figure}{Mini Batch $K$-Means Important Features}
\label{fig:mbkif}
\end{figure}

Figure \ref{fig:mbkif} shows the average variation of each features among different clusters. This shows that we can distinguish our clusters by considering different values for the features: county population, cumulative cases as of August 8, index of relative rurality, and number of ICU beds.

\subsection{Hierarchical Clustering} \label{sec:heirarchical} 

To find the best combination of linkage, distance metric, and number of clusters for hierarchical clustering, we trained a model with each different set of parameters on our dataset, computed its corresponding silhouette score, and returned a set of parameters with balance of a high score and a ``reasonable'' number of clusters. We note that the highest score always went to the clustering with as few clusters as possible. To counteract this, we chose the set of parameters that had the largest number of features that also had a silhouette score above 0.5.

Before running this grid search, we applied Principal Component Analysis (PCA) to our data to reduce the number of features in our dataset. PCA works by replacing the features in our dataset with new features that are linear combinations of the original features. It aims to capture the directions of greatest variance in the data, and reduce any redundancy across our feature space. Here, we reduced the original set of features down to 8 principal component features, which together capture 90\% of the variance we see in our data. Applying our grid search method to this feature-engineered dataset and following the process described in the previous paragraph, we chose the following set of parameters: average linkage (the distance between clusters is the average of all pairwise distances), Euclidean distance metric, and 25 clusters.

\begin{figure}[ht]
\centering
  \includegraphics[width=0.8\linewidth]{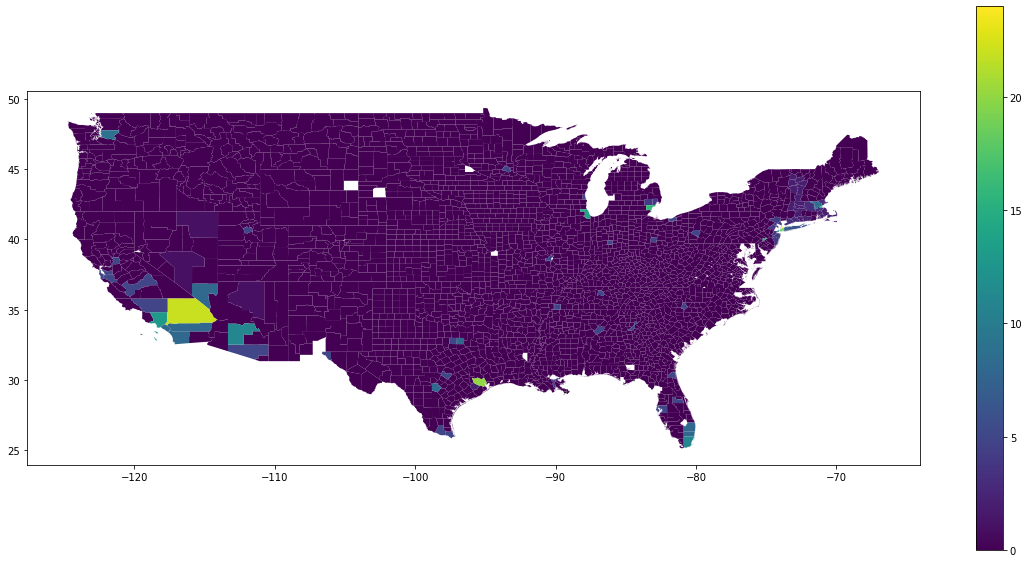}
  \caption{Map of US Counties, colored by hierarchical cluster.}
  \label{fig:agglommap}
\end{figure}

In Figure \ref{fig:agglommap} we see a graph of the counties in our dataset, excluding those in HI and AK, colored by their cluster number. As can be seen in the graph, there is one cluster, which we refer to as Cluster 0, that contains most of the counties. In particular, it contains 2971 counties, which is almost 99\% of all the counties in our dataset. Of the remaining 24 clusters, 13 contained only 1 county, and the remaining 11 clusters contained between 2 and 31 counties, for a total of 106 counties outside the one large cluster. 

In the following few sections, we present some insights from this hierarchical clustering approach. First, we describe the characteristics that distinguish counties in Cluster 0 from the remaining counties. Then, we discuss what distinguishes the clusters of counties outside Cluster 0 from each other. Finally, we present some comments on the overall results of hierarchical clustering applied to this dataset.

\subsubsection{Distinguishing Cluster 0 from All Other Clusters}

\begin{figure}[ht]
\centering
  \includegraphics[width=0.8\linewidth]{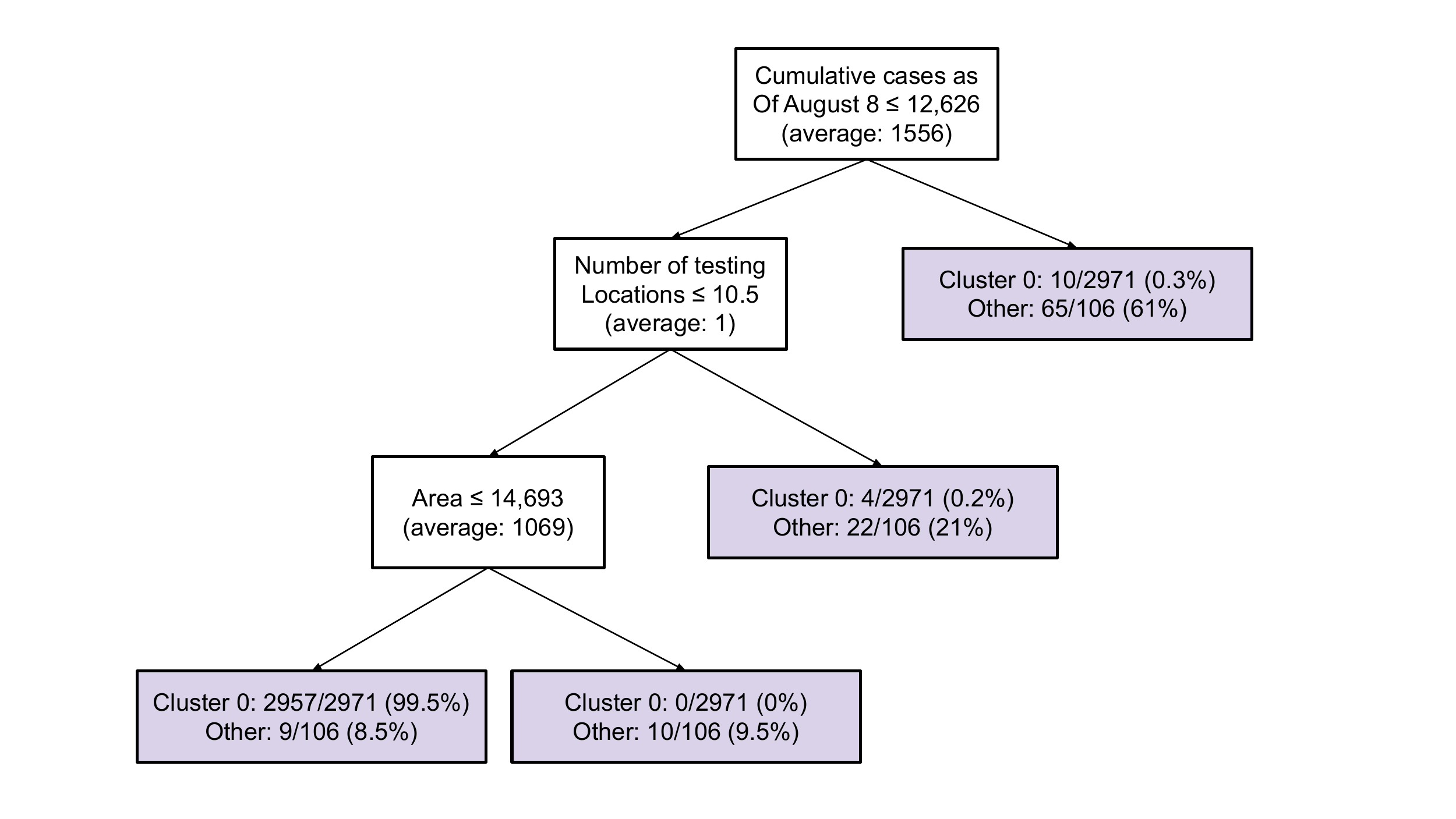}
  \caption{Simplified Decision Tree Distinguishing Cluster 0 from All Other Clusters.}
  \label{fig:cluster0velse}
\end{figure}

We first explore the differences between the counties in Cluster 0 and the counties outside that one cluster, by building a decision tree classifier to distinguish these two groups, shown in Figure \ref{fig:cluster0velse}. The first split in the resulting tree looks at whether a county has more or less than 12,626 cumulative COVID-19 cases as of August 8, which is more than 1.5 standard deviations above the average value of about 1600. Of the counties with more than 12,626 cumulative cases, we see that 65 of them are not in Cluster 0, which is about 61\% of the remaining counties, and 10 of them are in Cluster 0, which is only about 0.3\% of the counties in that cluster. Thus, we note that over half of the non-Cluster 0 counties have an incredibly high cumulative COVID-19 case count, and make up a majority of counties with such high cumulative case count.

Along the left branch of our decision tree, for which counties had fewer than 12,626 cumulative COVID-19 cases as of August 8, the next split is whether a county has more or less than 10 COVID-19 testing locations, which is about 3 standard deviations above the mean of 1. Of the counties with more than 10 testing locations, we have 22 counties outside Cluster 0, which makes up another 21\% of such counties, and we have 4 counties in Cluster 0, which is only about 0.2\% of Cluster 0 counties. Thus, we see that another subset of the non-Cluster 0 counties are those with an especially large number of testing locations.

The last split we point out from this tree is again along the left branch of the previous split, for which counties had fewer than 10 testing locations. This split looks at whether the area of the county is more or less than 14,693 square miles, which is more than 4 standard deviations above the mean of 1069 square miles. Of those counties whose area is more than 14,693 square miles, 10 were from outside Cluster 0, which is about 9.5\% of such counties, and none were from Cluster 0. Here, we see that another subset of the non-Cluster 0 counties have incredibly large areas, and in fact, this accounts for all the particularly large counties.

Of the counties along the branch of this split, for which counties have an area smaller than 14,693 square miles, 9 were from outside Cluster 0, which accounts for the remaining 8.5\% of such counties, and 2957 were from Cluster 0, which accounts for the remaining 99.5\% of these counties.

With these 3 splits of cumulative COVID-19 case count as of August 8, number of COVID-19 testing locations, and county area, we break off almost 87\% of non-Cluster 0 counties (97 counties) while only capturing about 0.4\% of Cluster 0 counties (14 counties). We also note that the value of each feature split represents some number significantly above the average value of that feature. This suggests that Cluster 0 might represent those counties whose features values are within a reasonable range as seen in the data, and those counties outside Cluster 0 represent outliers with respect to different features. This result could be due to the way that agglomerative clustering operates. Recall that it successively combines the nearest clusters, and so if one cluster starts to represent typical counties, then we would expect it to very often be the nearest cluster to some other county or cluster. This would explain why we see one particularly large cluster and then a few smaller clusters containing mostly outliers.

\subsubsection{Distinguishing the Smaller Clusters}
Now, we want to understand what distinguishes these remaining smaller clusters outside of Cluster 0, including the singleton clusters. We first note that the features with the highest variance among the remaining counties are, in order: (1) population (average outside Cluster 0: 1,121,476), (2) the number of cumulative COVID-19 cases as of August 8, which is the date on which we gathered the data (average outside Cluster 0: 22,595), (3) area in square miles (average outside Cluster 0: 4520), (4) number of new COVID-19 cases on July 23, which is around the date of the second US peak in cases (average outside Cluster 0: 3671), and (5) the number of cumulative COVID-19 deaths as of August 8 (average outside Cluster 0: 909). We will explore these features as we discuss the results that agglomerative clustering has generated.

\begin{figure}[ht]
\centering
  \includegraphics[width=0.8\linewidth]{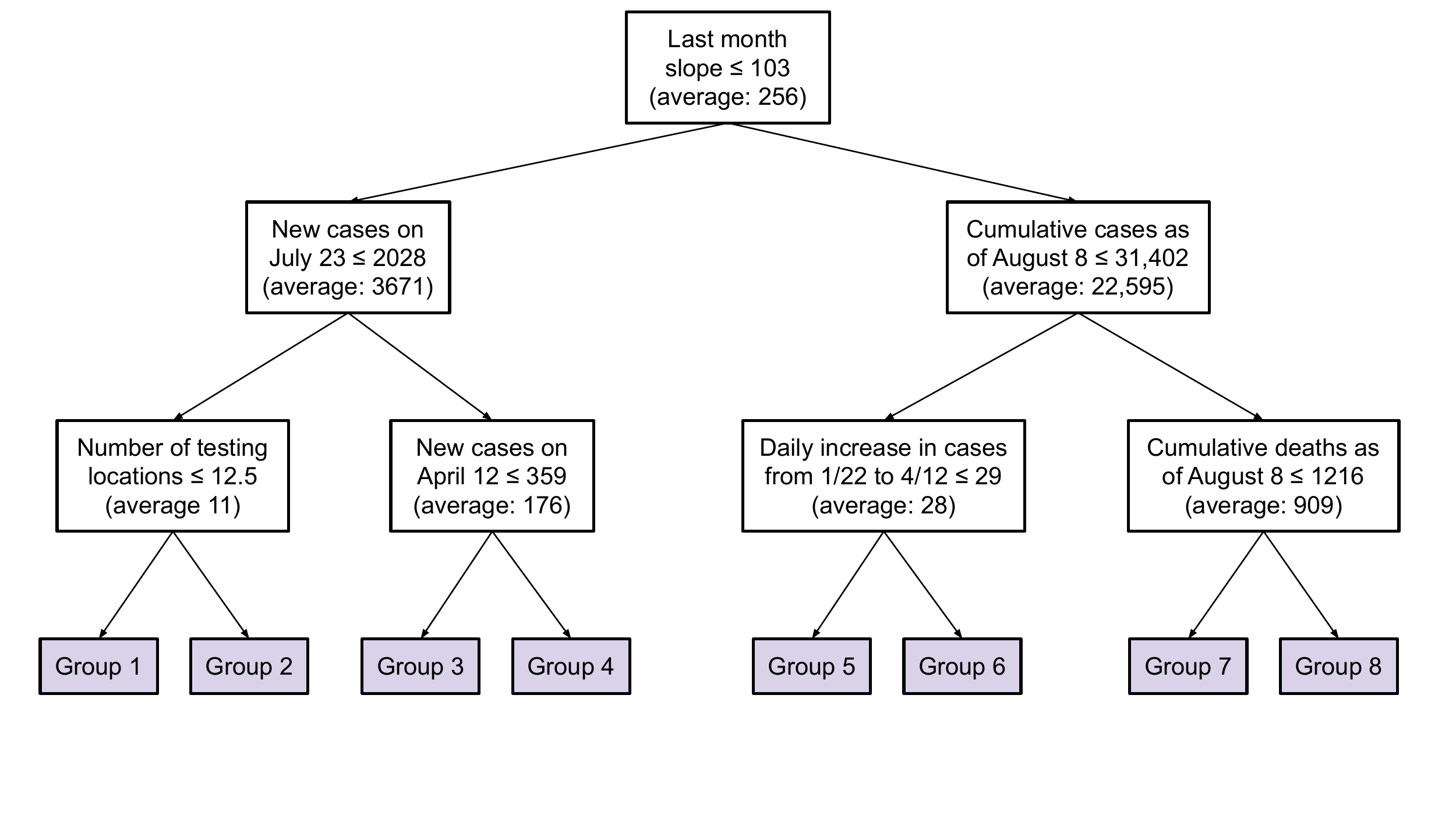}
  \caption{Simplified Decision Tree Distinguishing Groups of the Smaller Clusters.}
  \label{fig:smallclusters}
\end{figure}

In the agglomerative clustering process, we noticed that if we were to continue combining clusters, we would very often merge clusters with Cluster 0. Since many of our clusters contain only one or a few counties, in order to group them in a meaningful way for interpretation, we trained a decision tree to distinguish all clusters excluding Cluster 0. This allowed us to see which clusters landed in nearby leaf nodes, suggesting that they represent counties with similar feature values. In Figure \ref{fig:smallclusters} we show a simplified version of the resulting decision tree where we have grouped the clusters into 8 groups. We disregard clusters whose counties were split into a number of small groups across our decision tree, and focus only on clusters that remained completely or mostly intact within a leaf node. Across our 8 groups of clusters, we account for 22 of the remaining 24 clusters, which contain 102 of the remaining 106 counties.

Group 1 in our simplified decision tree contains 3 clusters, totalling 10 counties. Of these, seven are counties in Alaska, 2 are in Nevada, and one is in Arizona. With respect to the features that vary the most across non-Cluster 0 counties, the counties in Group 1 all have populations well below average; areas well above average; and new cases on July 23, total cases as of August 8, and total deaths as of August 8 well below average.

Group 2 contains 17 of the 18 counties from a single cluster and no other counties. The counties in this group are all in Massachusetts, Vermont, and Connecticut. Comparing the counties in this group to the general counties outside Cluster 0 with respect to the most variable features, we see that Group 2 has values for each (population, cumulative case count as of August 8, area, new cases on July 23, and cumulative death count as of August 8) that are below average. We split this group apart from Group 1 because the counties in Group 2 have a significantly smaller area than the counties in Group 1, and because for the remaining features, the counties in Group 1 seem to have much more extreme values.

Group 3 contains one cluster of size 5 and 11 counties from a cluster of size 12 for a total of 16 counties. The cluster of size 12 includes counties from across the country that all either include a larger city or are near a larger city, like Orleans, LA (New Orleans) and Macomb, MI (bordering Detroit). The cluster of size 5 contains a few counties in Massachusetts and a few in Connecticut. The counties in this group have populations below average, areas that are well below average, and total cases as of August 8 that are below average.

Group 4 contains a number of very small and singleton clusters, totalling 11 counties. These include 3 of the boroughs of New York City: Bronx, Queens, and Kings (Brooklyn). The other counties are all in New York (outside New York City) and New Jersey. The counties in this group all have well below average area, above average cases on July 23, and above average cumulative number of deaths as of August 8. Recall that New York City was hit really hard with COVID-19 cases at the beginning of the pandemic, and so it makes sense to see it as well as some of the surrounding counties grouped together.

Group 5 contains 28 of the counties from a cluster of size 31. The counties in this group are from all over the country and include Hennepin, MN (Minneapolis), Cuyahoga, OH (Cleveland), Milwaukee, WI, and Alameda, CA (Oakland). We note that the counties in this group have a population barely below average. They also have below average values for area, case count on July 23, total cases as of August 8, and total deaths as of August 8. Overall, this group represents counties with somewhat average feature values, and in fact, in the dendrogram for our hierarchical clustering, this group was fairly close to Cluster 0. This means that within a few more merges of agglomerative clustering, this group would have been combined with Cluster 0, which was the cluster representing counties with less extreme values.

Group 6 contains one singleton cluster. In particular, it contains the county of Wayne, MI (Detroit). We note that Wayne, MI has a population that is somewhat larger than average, and an area that is below average. We separated it from Group 5 because its values for case count on July 23, total cases as of August 8, and total deaths as of August 8 were all above or well above average. While this county may have non-COVID-19 related features with similar values to Group 5, for the features that describe COVID-19 case count and death count, Wayne, MI is an outlier.

Group 7 contains one cluster of size 8 and one singleton cluster, for 9 total counties. The singleton cluster is of San Bernardino, CA, and the other cluster contains counties that all contain a larger city, including Clark, NV (Las Vegas), San Diego, CA, and Dallas, TX. The counties in this group have larger than average populations, fewer than average cases on July 23 and fewer than average total deaths as of August 8, but more than average total cases as of August 8. Due to the fact that these counties have fewer cases in July and fewer deaths at the beginning of August but more cases at the beginning of August, this suggests that these counties may be at the beginning of a spike in cases.

Finally, Group 8 contains one cluster of size 2 and 3 singleton clusters, for 5 total counties. The singleton clusters contain the counties Harris, TX (Houston), Los Angeles, CA, and Cook, IL (Chicago). The cluster of size 2 contains Maricopa, AZ (Phoenix) and Miami-Dade, FL. This group differs from Group 7 in that the counties in Group 8 have significantly higher case counts on July 23 and significantly higher total deaths as of August 8. Additionally, the counties in Group 8 all have significantly higher than average populations,, and significantly higher total cases as of August 8. These counties represent some especially hard hit counties in the US, particularly over the summer.

\subsubsection{Summary of Hierarchical Clustering Results}
In our case, we see that agglomerative clustering creates one very large cluster that contains most of the counties in our dataset, and the remaining clusters tend to contain outliers with respect to some subset of our features. This could be due to the fact that we have relatively few features compared to the number of counties we have, and the values of these features are somewhat correlated, in particular for the COVID-19-related features. We attempted to mitigate this by applying PCA before clustering.

Changing the final total number of clusters in our agglomerative clustering mostly changed the number of clusters that were added to our one large cluster. Thus, after our initial cut off of 25 clusters, we further group together the remaining clusters using a decision tree. As we saw, these groups can be distinguished from each other based on their values for some subset of the 5 most variable features and by following the decision tree’s splits. The groupings we see in our tree seem fairly reasonable. For example, we see one group that contains most of the counties with the largest land area, another that contains counties in and around New York City, and another that contains many of the counties with the largest populations.

\subsection{OPTICS Clustering}
In this section we describe our work implementing the OPTICS clustering algorithm. We also discuss and interpret the clustering results.

\subsubsection{Dimension Reduction and OPTICS Clustering}
Before clustering our data, we applied the dimension reduction techniques PCA and t-SNE to the dataset. The PCA process is described above in Section \ref{sec:heirarchical} and t-SNE is similar, although it by definition reduces the dimensions of the dataset to either two or three dimensions. We generated two datasets with two and three attributes, respectively, using t-SNE. We also generated three datasets with three, five, and ten attributes by using PCA which store around 70\%, 85\% and 97\% of the variance in the original data, respectively.

For each of these five datasets, we applied the grid-search function we mentioned in the Section \ref{sec:optics1} to find the best set of parameters for the OPTICS clustering algorithm, as well as the best number of clusters to use given those parameters. The results of this grid-search returned higher values for both silhouette score and Calinski-Harabasz index for a large number of clusters for the datasets that we applied t-SNE to but returned higher values for both scores for fewer than five clusters for the datasets that we applied PCA to. Since we expected that ending up with fewer than five clusters would not produce especially useful results, we set conditions on our grid search to account for this and selected the resulting best combination of the values of parameters for each of these datasets, which we show in Table \ref{OPTICS_gridsearch}. 
\begin{table} 
    \scriptsize
    \begin{spacing}{1}
\renewcommand{\arraystretch}{1.3}
\begin{tabular}{|m{1.1cm}<{\centering}|m{1.8cm}<{\centering}|m{1.1cm}<{\centering}|m{1.4cm}<{\centering}|m{1.2cm}<{\centering}|m{1.4cm}<{\centering}|m{1.5cm}<{\centering}|m{1.1cm}<{\centering}|m{1.4cm}<{\centering}|}
		\hline 
		&
		Number of dimensions &
		Min samples&
		Perplexity&
		Number of clusters&
		Silhouette score&
		Calinski-Harabasz score&
		Model matrix&
		Silhouette matrix\\
		\hline
		& & & & & & & & \\[-1pt] 
		t-SNE&2&11&40&132&0.4101&59.5546&cosine&cosine\\
		& & & & & & & & \\[-0.5pt] 
		t-SNE&3&14&5&68&0.0080&57.2414&cosine&cosine\\
		& & & & & & & & \\[-0.5pt] 
		PCA&3&9&/&75&-0.3464&31.5931&cosine&cosine\\
		& & & & & & & & \\[-0.5pt] 
		PCA&5&9&/&32&-0.6436&78.9359&cosine&cosine\\
		& & & & & & & & \\[-0.5pt] 
		PCA&10&19&/&6&-0.2871&269.5029&cosine&cosine\\
		& & & & & & & & \\[-1pt] 
		\hline
	\end{tabular}
         \end{spacing}
	 \caption{\centering The Value of Parameters Selected from the Dataset after Applying Dimension Reduction Techniques}
	\label{OPTICS_gridsearch}
\end{table}

Based on the silhouette score and Calinski-Harabasz index, as well as the fact that this results in a fairly reasonable number of clusters, we decided to focus on the dataset to which we applied PCA in order to reduce to 5 features. In addition, we chose this dataset since with 5 principal components, we are able to explain 85\% of the variance in the original data.

\begin{figure}[ht]
    \centering
    \includegraphics[scale = 0.35]{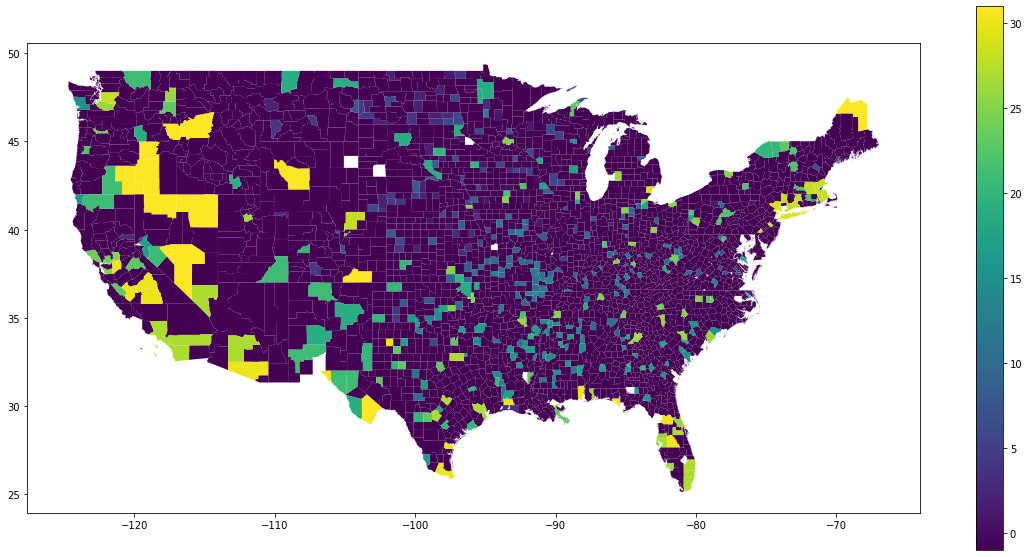}
    \caption{Location of the Counties in Each clusters \label{OPTICS_clustering}}
    \label{fig:my_label}
\end{figure}

The distribution of counties per cluster for this model is shown in Table \ref{distribution_of_clusters}, and a map with all counties, excluding those HI or AK colored by cluster number is shown Figure \ref{fig:my_label}. 

\begin{table}[htbp] \centering
    \scriptsize
    \begin{spacing}{1}
	    \renewcommand{\arraystretch}{1.3}
	    \begin{tabular}{|m{1.2cm}<{\centering}|m{2.1cm}<{\centering}|m{1.2cm}<{\centering}|m{2.1cm}<{\centering}|m{1.2cm}<{\centering}|m{2.1cm}<{\centering}|}
    		\hline
    		Labels&
    		Number of Counties&
    		Labels&
    		Number of Counties&
    		Labels&
    		Number of Counties
    		\\
    		\hline
    		-1&2579&10&40&21&10\\
    		0&15&11&14&22&16\\
    		1&19&12&19&23&13\\
    		2&14&13&16&24&10\\
    		3&14&14&9&25&21\\
    		4&15&15&16&26&16\\
    		5&12&16&14&27&17\\
    		6&21&17&13&28&14\\
    		7&11&18&13&29&17\\
    		8&12&19&19&30&14\\
    		9&16&20&21&31&19\\
    		\hline
    	\end{tabular}
    	\end{spacing}
	    \caption{\centering Distribution of Clusters \label{distribution_of_clusters}} 
	\label{table1}
\end{table}

We can see the majority of counties are labeled as noise, corresponding to cluster -1, shown in Figure \ref{OPTICS_clustering}. This can be explained by the fact that the most of the confirmed cases and death are concentrated in a small number of counties while the values for these features in most of the other counties are extremely small. This causes large distances between these counties in the dataset. Thus, the counties where these values are small are prevented from being assigned a cluster and are labeled as outliers.

\subsubsection{Summary of OPTICS Results}
\begin{figure}[ht]
    \centering
    \includegraphics[scale = 0.3]{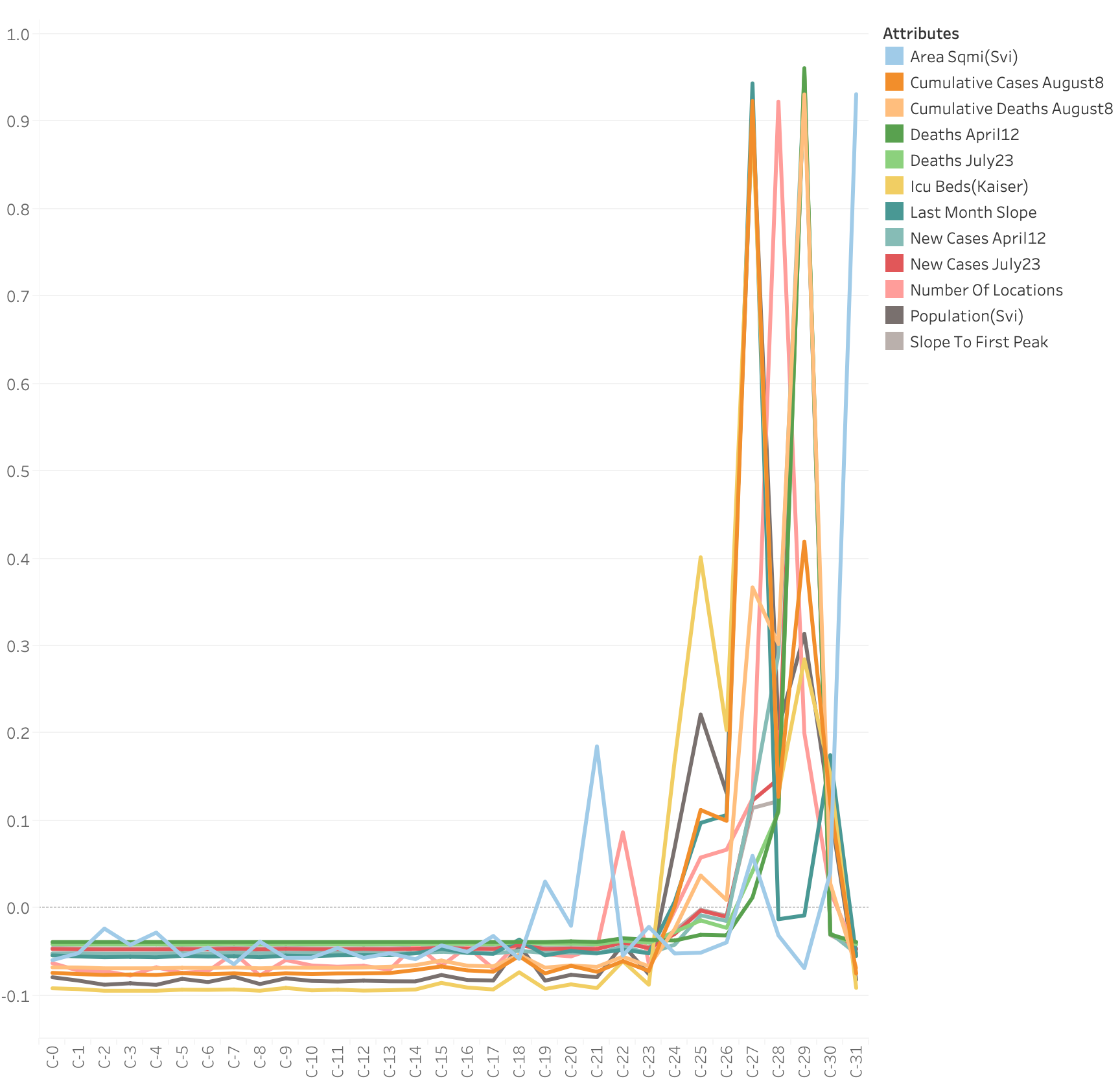}
    \caption{\centering The Mean Value of Each Attribute with Clusters \label{mean_value_of_attributes} }
\end{figure}

After we removed the outliers, which correspond to counties in cluster -1, we wanted to see how the values of the features are similar or different across the remaining clusters. To do this, we calculated the standard deviation for each attribute and removed those with relatively small standard deviations since they represent features with similar values across all clusters. Then for each cluster, we computed the mean value of the 12 attributes in our data with larger standard deviations and plotted these values along with the cluster name in Figure \ref{mean_value_of_attributes}.

We can see from Figure \ref{mean_value_of_attributes} that for clusters 0 through 23, as well as for cluster 31, the values of most attributes are fairly constant and also are slightly negative. On the other hand, clusters 24 to 30 have extremely large value of each attributes compared with other clusters. Of the 12 attributes that we have plotted in Figure \ref{mean_value_of_attributes}, 11 have their highest average value in cluster 27, cluster 28 and cluster 29. In the discussion below, we will analyze these three clusters to figure out what caused their unusual values.
   
\begin{table}[htbp] \centering
    \begin{minipage}[b]{0.3\linewidth}
    \footnotesize
    \begin{spacing}{1}
	\renewcommand{\arraystretch}{1.2}
	    \begin{tabular}{|m{1.4cm}<{\centering}|m{2.5cm}<{\centering}|}
    		\hline
    		\textbf{Cluster} &
    		\textbf{Location}
    		\\
    		\hline
    		27&Hillsborough, FL\\
    		27&Palm Beach, FL\\
    		27&Maricopa, AZ\\
    		27&San Diego, CA\\
    		27&Broward, FL\\
    		27&Orange, FL\\
    		27&Travis, TX\\
    		27&Los Angeles, CA\\
    		27&Orange, CA\\
    		27&Riverside, CA\\
    		27&Gwinnett, GA\\
    		27&Clark, NV\\
    		27&Tarrant, TX\\
    		27&Miami-Dade, FL\\
    		27&Bexar, TX\\
    		27&Harris, TX\\
    		27&Dallas, TX\\
    		\hline
	    \end{tabular}
	\end{spacing}
	\caption{\centering County List of Cluster 27 \label{tab:cluster27}}
	\end{minipage}\hfill
\begin{minipage}[b]{0.65\linewidth}
\centering
\includegraphics[width=1\linewidth]{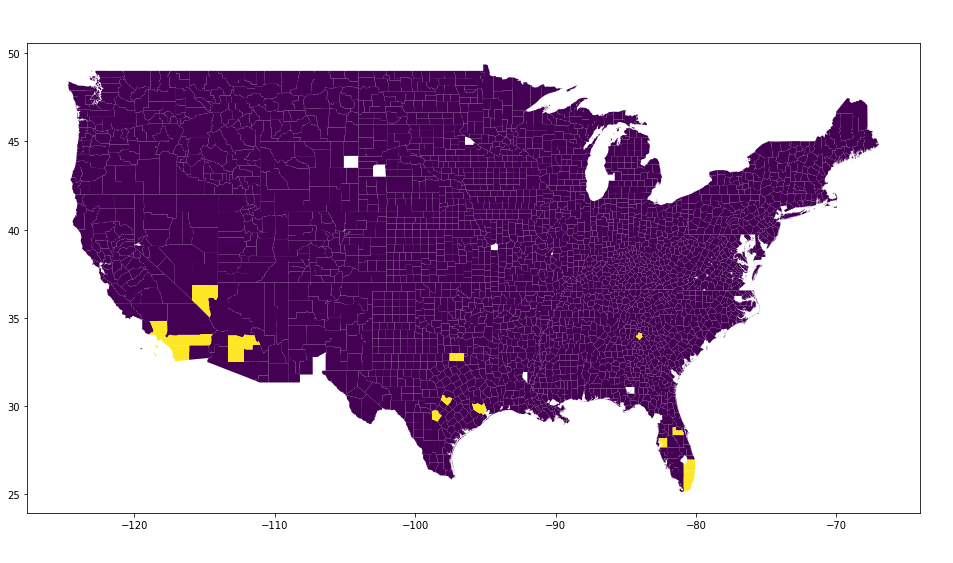}
\captionof{figure}{County Location of Cluster 27 \label{fig:cluster27}}
\end{minipage}
\end{table}
\bigskip

Cluster 27 contains 17 counties located in 6 different states. We have listed these counties in Table \ref{tab:cluster27} and shown them on a map in Figure \ref{fig:cluster27}. Most of the counties in this cluster are concentrated at California, Texas and Florida and are located in the southern part of the US. This cluster has the highest mean value of area, population, number of ICU beds and number of cumulative cases as of August 8. Also, it has an extremely high mean value of the case growth rate for the month leading up to August 8, compared with the rest of the clusters. Additionally, the majority of these counties have high populations compared to other counties in their state. For example, Harris, Dallas, Tarrant, Bexar and Travis are the top 5 counties in population in Texas. We know that the states of California, Texas, and Florida all have high numbers of confirmed cases after June 2020. With the above information, we could conclude that this cluster contains the counties with most rapid growth of the number of confirmed cases and deaths after June.

\begin{table}[htbp] \centering
    \begin{minipage}[b]{0.3\linewidth}
    \footnotesize
    \begin{spacing}{1.2}
	\renewcommand{\arraystretch}{1.2}
	    \begin{tabular}{|m{1.4cm}<{\centering}|m{2.5cm}<{\centering}|}
    		\hline
    		\textbf{Cluster} &
    		\textbf{Location}
    		\\
    		\hline
    		28&Plymouth, MA\\
    		28&Worcester, MA\\
    		28&Dutchess, NY\\
    		28&Weld, CO\\
    		28&Bristol, MA\\
    		28&Norfolk, MA\\
    		28&Fairfield, CT\\
    		28&Hartford, CT\\
    		28&New Haven, CT\\
    		28&Middlesex, MA\\
    		28&Delaware, PA\\
    		28&King, WA\\
    		28&Essex, MA\\
    		28&Mercer, NJ\\
    		\hline
	    \end{tabular}
	\end{spacing}
	\caption{\centering County List of Cluster 28 \label{tab:cluster28}}
	\end{minipage}\hfill
\begin{minipage}[b]{0.65\linewidth}
\centering
\includegraphics[width=1\linewidth]{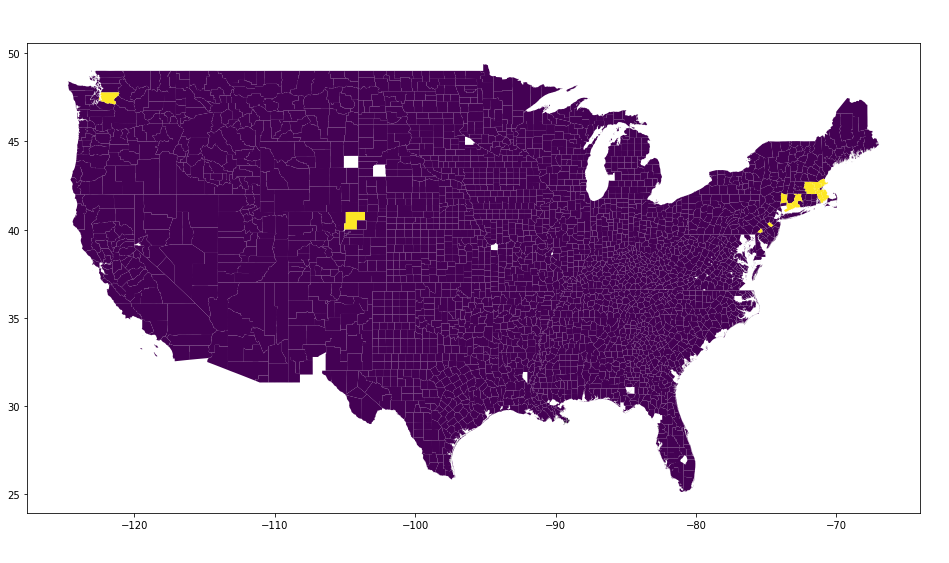}
\captionof{figure}{County Location of Cluster 28 \label{fig:cluster28}}
\end{minipage}
\end{table}
\bigskip

Cluster 28 contains 14 counties located in 7 different states. We have listed these counties in Table \ref{tab:cluster28} and shown them on a map in Figure \ref{fig:cluster28}. We can see that most of these counties are in  Massachusetts, Connecticut, New York, New Jersey, and Pennsylvania. Almost all the counties are located in the northeast part of the US except King, WA and Weld, CO. This cluster has an average population of 799,174, has the highest average number of testing locations, and has a relatively high nursing home population. This indicates the counties in this cluster have good medical resources and focused on stopping the spread of COVID-19 by adding more testing locations. This cluster also contain counties with large populations and relatively low cumulative confirmed cases as of August 8 compared with other clusters.

\begin{table}[htbp] \centering
    \begin{minipage}[b]{0.3\linewidth}
    \footnotesize
    \begin{spacing}{1.2}
	\renewcommand{\arraystretch}{1.2}
	    \begin{tabular}{|m{1.3cm}<{\centering}|m{2.5cm}<{\centering}|}
    		\hline
    		\textbf{Cluster} &
    		\textbf{Location}
    		\\
    		\hline
    		29&Suffolk, NY\\
    		29&Nassau, NY\\
    		29&Orange, NY\\
    		29&Bergen, NJ\\
    		29&Richmond, NY\\
    		29&Westchester, NY\\
    		29&Orleans, LA\\
    		29&Wayne, MI\\
    		29&New York, NY\\
    		29&Hudson, NJ\\
    		29&Passaic, NJ\\
    		29&Union, NJ\\
    		29&Kings, NY\\
    		29&Queens, NY\\
    		29&Rockland, NY\\
    		29&Essex, NJ\\
    		29&Bronx, NY\\
    		\hline
	    \end{tabular}
	\end{spacing}
	\caption{\centering County List of Cluster 29 \label{tab:cluster29}}
	\end{minipage}\hfill
\begin{minipage}[b]{0.65\linewidth}
\centering
\includegraphics[width=1\linewidth]{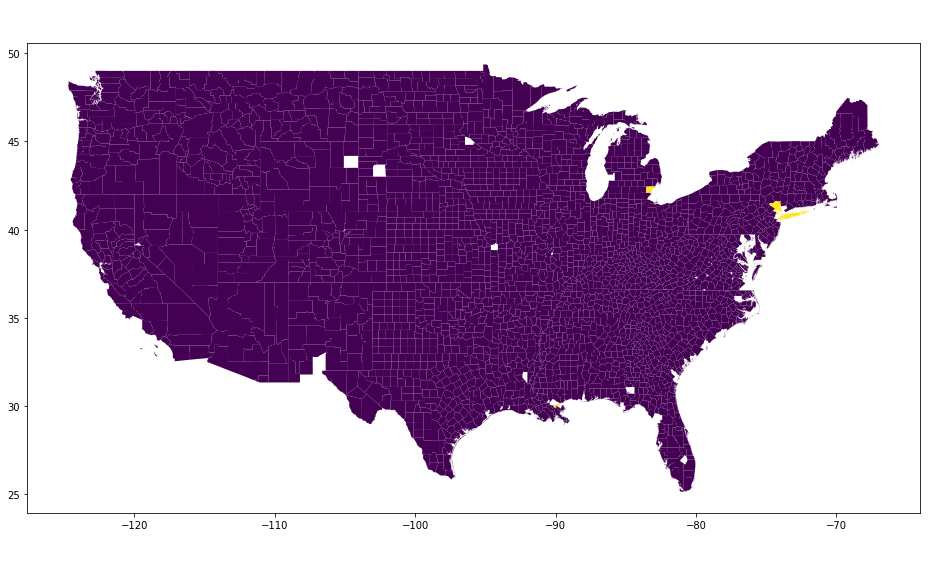}
\captionof{figure}{County Location of Cluster 29 }
\label{fig:cluster29}
\end{minipage}
\end{table}

Cluster 29 contains 17 counties located in 4 different states. We have listed these counties in Table \ref{tab:cluster29} and shown them on a map in Figure \ref{fig:cluster29}. Fifteen of these counties are in New York and New Jersey. This cluster has the highest average number of new cases on April 12, the highest average number of cumulative deaths as of August 8, and the highest average growth rate for the first month of the pandemic. Based on this information, we see that this cluster grouped the counties with the highest number of confirmed cases and deaths before June. This agrees with the fact that when COVID-19 first started to spread in the US, New York and New Jersey had particularly high numbers of cases. 


\section{Conclusion}
Among the prototype-based methods, we saw that $K$-means and Fuzzy-$c$ means clustering methods produced somewhat similar results, which makes sense since they are similar algorithms. Both approaches clustered many counties in the southern half of the country with counties in the west into one cluster, counties in the northern half of the country into another cluster, and some seemingly outlier counties into another few clusters. The Gaussian mixture model and Mini Batch $K$-means produced similar clusters to each other. Both had one very large cluster of counties and another cluster or two containing far fewer counties.

Across these four prototype-based methods, we saw almost all of our features were roughly equally as useful in distinguishing our clusters, suggesting that all of our features could provide useful insight into our data. According to the important features selected by each of the prototype-based approaches, we saw that a county's population, index of relative rurality, number of ICU beds, and ranking with respect to socioeconomic status came up most frequently. Thus, we conclude that these features are closely related to how severely the COVID-19 pandemic impacted those counties. The features describing the county ranking with respect to housing and transportation as well as the county ranking with respect to household decomposition and disability also seem related to the severity of the COVID-19 pandemic in a given county to some extent.

The hierarchical clustering and the density-based clustering via OPTICS both output one very large cluster of counties that we can view as ``noise'' or counties with more average values for each of our features, and a number of much smaller clusters. In hierarchical clustering, we found a number of singleton clusters, and only a few clusters with sizes in between. Due to the nature of the OPTICS algorithm, the smaller clusters all contained at least a pre-specified number of counties, resulting in more uniform sizes for these smaller clusters. Since we were ultimately interested in finding similar counties, this suggests that perhaps the OPTICS algorithm is better suited between these two methods.

Overall, from this project, we found that applying clustering to US county data related to the COVID-19 pandemic revealed a variety of insights. Due to the nature of clustering, we did not select one best approach, and instead found that each technique revealed different patterns and structure within our data. 

\newpage

\bibliography{\jobname} 
\end{document}